\documentclass[a4paper,11pt]{article}
\usepackage[top=0.75in, bottom=0.75in, left=0.85in, right=0.85in]{geometry}
\usepackage{graphicx}
\usepackage{url}
\usepackage{palatino}
\usepackage{tabularx}
\fontfamily{SansSerif}
\selectfont
\usepackage[
    colorlinks=true,       
    linkcolor=blue,        
    citecolor=green,       
    urlcolor=red           
]{hyperref}
\usepackage{amsmath}  
\usepackage{amssymb}  
\usepackage{tikz}
\usepackage{comment}
\usepackage{subcaption}
\usepackage{float}
\usepackage{authblk}
\usepackage{amsmath}
\usepackage{cite}
\usepackage{subcaption}
\usepackage{bm}
\usepackage[table,dvipsnames]{xcolor}

\usepackage[T1]{fontenc}
\usepackage
[utf8]
{inputenc}

\definecolor{mygrey}{gray}{0.75}
\newcommand{\dt}[1]{\frac{\text{d}#1}{\text{d}t}}
\newcommand{\dtt}[1]{\frac{\text{d}^2#1}{\text{d}t^2}}

\title{Crack Path Prediction with Operator Learning using Discrete Particle System data Generation}
\date{}
\author[1*]{Elham Kiyani}
\author[2*]{Venkatesh Ananchaperumal}
\author[1]{Ahmad Peyvan}
\author[3]{Mahendaran Uchimali}
\author[2]{Gang Li}
\author[1]{George Em Karniadakis}
\affil[1]{Division of Applied Mathematics, Brown University, USA}
\affil[2]{Department of Mechanical Engineering, Clemson University, USA}
\affil[3]{School of Mechanical Sciences, Indian Institute of Technology Bhubaneswar, India}
\affil[*]{Authors contributed equally to this work.}

\begin{document}
\maketitle
\begin{abstract}
Accurately modeling crack propagation is critical for predicting failure in engineering materials and structures, where small cracks can rapidly evolve and cause catastrophic damage. The interaction of cracks with discontinuities, such as holes, significantly affects crack deflection and arrest. Recent developments in discrete particle systems with multibody interactions based on constitutive behavior have demonstrated the ability to capture crack nucleation and evolution without relying on continuum assumptions. In this work, we use data from Constitutively Informed Particle Dynamics (CPD) simulations to train operator learning models, specifically Deep Operator Networks (DeepONets), which learn mappings between function spaces instead of finite-dimensional vectors. We explore two DeepONet variants: vanilla and Fusion DeepONet, for predicting time-evolving crack propagation in specimens with varying geometries. Three representative cases are studied: (i) varying notch height without active fracture; and (ii) and (iii) combinations of notch height and hole radius where dynamic fracture occurs on irregular discrete meshes. The models are trained using geometric inputs in the branch network and spatial-temporal coordinates in the trunk network. Results show that Fusion DeepONet consistently outperforms the vanilla variant, with more accurate predictions especially in non-fracturing cases. Fracture-driven scenarios involving displacement and crack evolution remain more challenging. These findings highlight the potential of Fusion DeepONet to generalize across complex, geometry-varying, and time-dependent crack propagation phenomena.
\end{abstract}

\section{Introduction}
Brittle fracture has long been a critical topic in the study of material failure, with significant implications in structural engineering, materials science, and geomechanics. It refers to a fracture process that occurs with minimal plastic deformation and is often accompanied by rapid crack propagation \cite{bouchbinder2014dynamics}. Due to its sudden and catastrophic nature, understanding the mechanisms of brittle fracture is essential for designing safer and more reliable materials and structures \cite{wahab2023review}.

Over the decades, experimental investigations have played a pivotal role in characterising the fracture behaviour of brittle materials such as ceramics, glasses, rocks, and certain metals under specific conditions. A wide range of techniques has been employed to determine fracture toughness, analyse crack paths, and explore surface features and underlying microstructural failure mechanisms. Methods such as Scanning Electron Microscopy (SEM), Digital Image Correlation (DIC), acoustic emission monitoring, and X-ray computed tomography have enabled researchers to probe the initiation and evolution of cracks at various length scales. These techniques provide insights into the influence of microstructure, flaws, and loading conditions on brittle crack propagation.

Parallel to experimental efforts, the computational modelling of brittle fracture has been extensively studied to predict crack initiation, growth, and interaction under various conditions. Several numerical techniques have been developed to describe crack paths based on different failure criteria and material models. Among them, the Extended Finite Element Method (XFEM)~\cite{bely_xfem}, Finite Fracture Mechanics (FFM)~\cite{taylor_fem_fracture}, and the Embedded Finite Element Method (EFEM)~\cite{ortiz_efem} have emerged as prominent approaches for simulating brittle crack propagation. XFEM, for instance, allows discontinuities to be represented independently of the mesh by enriching the displacement field with special functions, enabling the simulation of arbitrary crack growth without remeshing. FFM introduces an energy-based criterion for crack advance, taking into account both the energy release rate and fracture resistance. EFEM, on the other hand, modifies the finite element formulation to embed a crack within an element, providing an alternative to mesh refinement around the crack tip. In parallel, the Cohesive Zone Model (CZM) has been widely adopted to simulate fracture by incorporating a traction-separation law that governs damage evolution ahead of the crack tip \cite{dugdale1960yielding_cohesive}.

Despite their advantages, these computational methods come with significant challenges and limitations. One major difficulty lies in the accurate tracking of the crack path evolution, especially in heterogeneous or complex materials. For methods like XFEM and EFEM, a splitting algorithm must be defined for elements that are either intersected by the crack or contain the crack tip. This becomes computationally demanding and can result in crack topology problems, particularly when quadratic or higher-order displacement interpolation schemes are used \cite{diehl2022comparative}. 

Regularised continuum models such as the phase field model~\cite{steinbach2009phase,kiyani2023framework,moelans2008introduction,wu2020phase,kiyani2022machine,schreiber2020phase} are popularly used to describe the evolution of cracks~\cite{luo2023phase,lorenzis2020numerical,bourdin2000numerical,bourdin2008variational,choo2018coupled,heider2020phase}. These regularised models incorporate higher gradients of the field variables, resulting in diffuse interfaces. Notably, the implicit kinetic relations of phase field models do not account for the micro-scale effects. Most importantly, the interface thickness determines the length scale (in the range of nm) of the model, which provides additional limitations. Another continuum model adopted to describe the crack initiation and propagation is peridynamics, where the horizon radius plays a crucial role in the model result.

Meanwhile, mesoscale discrete models are shown to be capable of describing the micro-scale features and their influence on the macroscopic behaviour. The fundamental difference of discrete models from their continuum counterparts lies in treating the body as a discrete or continuum entity. The consequence of this assumption is discussed in many aspects; specifically, the dispersion relation for an elastic material is linear in continuum, whereas the corresponding one for discrete models is sinusoidal. Thus, another limiting feature of the continuum approach is the non-dispersive nature of the model. On the other hand, discrete models, even with the simplest particle interactions, are dispersive. Thus, in discrete models, the lattice-level dispersive effects arise naturally, and these models are naturally suitable for describing cracks or any microscale phenomenon \cite{abeyaratne2003lattice, kalia2003multiresolution}.

Recent advances in discrete particle dynamics have enabled the internal forces among particles to be derived directly from a material’s continuum free energy using multi-body interactions \cite{CPD_CMAME, CPD_Acta, CPD_IJMS, CPD_hyper_CPM, CPD_composites_MAMS}. This Constitutively informed Particle Dynamics (CPD) formulation overcomes the limitations of traditional pairwise lattice-based models by allowing the complete elastic behavior of a material to emerge from its constitutive matrix or strain energy function, rather than relying on predefined network architectures. The initial configuration of the particle system is established through Delaunay triangulation, which serves as a fixed reference lattice for computing internal force resistances. By capturing the interactions and displacements between particles, the CPD method facilitates accurate computation of stress, strain, and failure behavior in heterogeneous materials \cite{CPD_CMAME, CPD_Acta}.

While CPD provides a physically grounded framework for simulating complex deformation and fracture processes, it can be computationally expensive when applied to large-scale systems or across varying geometries. In parallel, machine learning has recently emerged as a powerful suite of tools for modeling complex material behavior, including fracture~\cite{fuhg2024review,raissi2019physics,manav2024phase,kiyani2024predicting,goswami2020transfer}. These approaches have been successfully applied to phase-field fracture modeling~\cite{manav2024phase,kiyani2024predicting}, damage evolution in quasi-brittle materials~\cite{zheng2022physics}, and fatigue crack growth prediction~\cite{chen2024crack}, offering data-driven surrogates that can significantly accelerate predictive simulations.

Machine learning models have increasingly been adopted to simulate particle motion in discrete systems, offering significant computational advantages over traditional Discrete Element Method (DEM) simulations. For example, a Convolutional Neural Network (CNN) has been employed to update particle positions using intermediate velocities and learned correction terms~\cite{lu2021machine}. In powder mixing applications, a Recurrent Neural Network with Stochastic Randomness (RNNSR) captures both Lagrangian and Eulerian components of particle behavior to predict individual trajectories~\cite{li2014advances}. The NN4DEM framework introduces a physics-informed CNN by embedding contact force models directly into convolutional kernels, enabling structured-grid predictions that retain physical fidelity~\cite{naderi2024discrete}. In contrast, Kazemi et al.~\cite{kazemi2025novel} proposed an ML-DEM approach that couples short-run DEM data with a deep network employing continuous convolution operators to forecast particle displacements in rotary drums with high accuracy and reduced cost.

Beyond surrogate modeling, operator learning has emerged as a powerful paradigm in scientific machine learning, particularly for approximating solutions to partial differential equations (PDEs). Unlike conventional regression-based models, operator learning seeks to approximate mappings between infinite-dimensional function spaces, offering generalization across varying initial/boundary conditions and geometries~\cite{lu2019deeponet,lu2024bridging,shih2024transformers,kovachki2023neural,CiCP-35-1194,khodakarami2025mitigating,peyvan2025fusion}. A leading architecture in this domain is the Deep Operator Network (DeepONet)\cite{lu2019deeponet}, which uses a branch-trunk structure to separately encode input functions and query locations. Physics-informed DeepONets extend this framework by incorporating residuals of governing equations into the loss function\cite{wang2021learning,goswami2020transfer,kiyani2024predicting}.
%
Further enhancements in training efficiency and model generalization have been achieved through two-step training procedures\cite{lee2024training,peyvan2024riemannonets,kiyani2024predicting}, which decouple the optimization of the branch and trunk networks before recombining them. This strategy was recently applied in the context of brittle fracture modeling using phase-field formulations\cite{kiyani2024predicting}, where various trunk architectures—including standard feedforward networks, Kolmogorov–Arnold Networks (KANs)~\cite{liu2024kan,sprecher2002space,koppen2002training}, and physics-informed designs—were evaluated for their ability to accurately capture crack nucleation, propagation, and branching behaviors.
In parallel, Crack-Net~\cite{xu2025crack} introduces a specialized deep learning framework for predicting crack propagation and stress–strain curves in particulate composite materials. By incorporating an implicit constraint that links the evolution of cracks to mechanical responses within the network architecture, Crack-Net enables data-efficient, long-term predictions of fracture behavior across diverse material morphologies and interface conditions. To address challenges related to complex geometries and irregular computational grids, the Fusion DeepONet framework~\cite{peyvan2025fusion} further extends the classical DeepONet by introducing a multi-scale conditioned neural field. This allows for accurate and geometry-aware predictions across heterogeneous domains, making it particularly suitable for modeling responses in evolving or non-uniform materials where spatial variations and structural features play a critical role.

Building on this foundation, the present work integrates DeepONets with the CPD framework to predict particle-level displacements and fracture evolution. This data-driven methodology combines the physical interpretability of CPD with the expressive representational capacity of neural operators, enabling accurate and efficient prediction of crack propagation across varying material configurations without the need for retraining on each new geometry.

In this study, we evaluate and compare the performance of vanilla DeepONet and Fusion DeepONet architectures for predicting crack evolution over 100 time steps. To assess the models’ generalization and robustness, we design three representative scenarios:
\begin{enumerate}
    \item Varying the pre-crack notch height above a hole with fixed radius ($r = 1~\text{cm}$),
    \item Repeating the first scenario while explicitly modeling failed elements in the domain, and
    \item Varying the hole radius while maintaining a fixed notch position located $1.5~\text{cm}$ above the hole center.
\end{enumerate}

These cases serve to test the models' ability to generalize across diverse geometric and failure configurations inherent to fracture dynamics in complex materials.

The paper is structured as follows:
Section~\ref{sec:Constitutively_informed_Particle_Dynamics} provides a comprehensive review of the Constitutively informed Particle Dynamics (CPD) framework and its general formulation. In Section~\ref{sec:Modeling geometric discontinuity}, we discuss the modeling of geometric discontinuities and brittle cracks, as well as their influence on crack path evolution. Section~\ref{sec:Operator_Learning} introduces the DeepONet and Fusion DeepONet architectures, along with the details of the training process and network design. In Section~\ref{sec:Performance of DeepONet and Fusion-DeepONet}, we present an analysis of the performance of the vanilla and Fusion DeepONet models for data-driven crack propagation prediction, and the paper concludes with a summary in Section~\ref{sec:Summary}.

\section{ Constitutively informed Particle Dynamics }\label{sec:Constitutively_informed_Particle_Dynamics}
The CPD model introduces a way for the discrete system to describe a general constitutive behavior through the definition of interparticle forces via continuum free energy~\cite{CPD_CMAME}. Multiple forms of free energy have been used with CPD, such as non-convex free energy for solid phase transformations~\cite{CPD_Acta, CPD_IJMS}, neo-Hookean free energy for hyperelastic behaviours~\cite{CPD_hyper_CPM}, and combinations of these with isotropic elastic free energy to describe heterogeneous composites~\cite{CPD_composites_MAMS,CPD_composites_SMA}. This section provides a brief overview of the formulation. Detailed descriptions of the model and its applications for characterizing evolving cracks and microstructure can be found in \cite{CPD_CMAME}.

\begin{figure}[!tbh]
    \centering
    {\includegraphics[width=0.7\textwidth, trim= 0cm 10cm 0 0, clip]{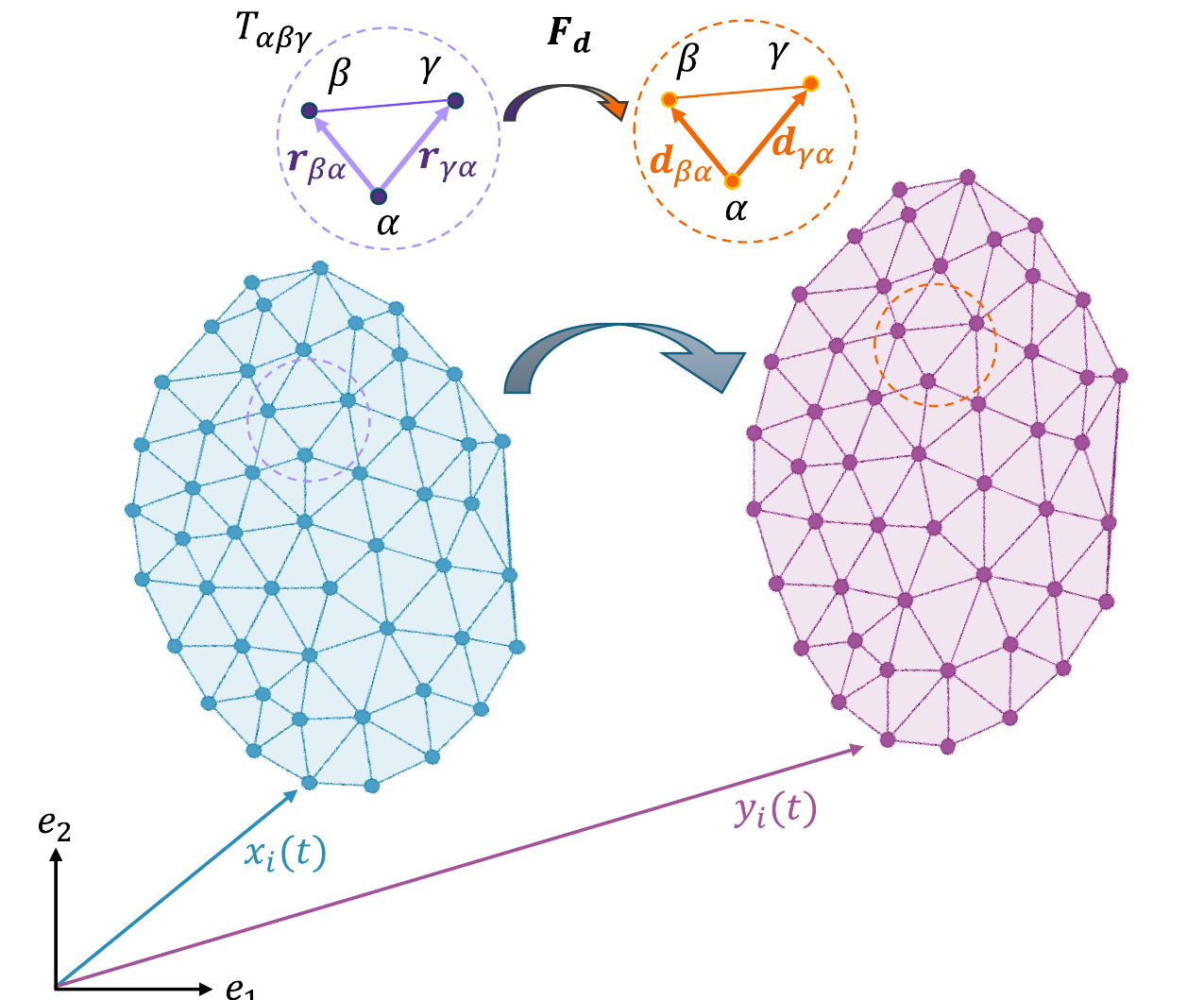}} 
    \caption{Discretization and mapping of deformation of the domain through particle displacement.}
    \label{fig:Potato}
\end{figure}

The domain is treated as a collection of irregularly distributed interacting particles. The model describes the behaviour of solid materials through the three particle interactions using the reference configuration ${\bm x}_i (t)$ and the deformed configuration represented as ${\bm y}_i (t)$ as shown in Figure \ref{fig:Potato}. By defining the inter-particle forces as a function of ${\bm x}_i (t)$, ${\bm y}_i (t)$ and material properties through continuum free energy, the current trajectory of the particle ${\bm y}_i (t)$ is obtained by solving Newton's equations of motion for every particle. The governing equations are given,
\begin{equation}
m_i \dtt{{\bm y}_i} + c\dt{{\bm y}_i} = {\bm f_i}, \quad i=1,2,\dots,N,
\label{eq:eom}
\end{equation}
where $m_i$ denotes the mass of particle $i$, and $c$ represents a damping coefficient. The net force ${\bm f}_i$ on particle $i$, results from interacting neighbor particles. The interacting neighbors for a particle are identified through Delaunay triangulation in reference configuration, as shown in Figure~\ref{fig:Potato}.

The specific triangle ${\cal T}_{\alpha\beta\gamma}$ composed of particles $\alpha, \beta$, and $\gamma$ in reference and deformed configurations is shown as the inset in Figure~\ref{fig:Potato} to demonstrate the methodology to obtain multi-body interaction. The deformation of the triangle ${\cal T}_{\alpha\beta\gamma}$ described by relative position vectors in reference and current configurations, respectively, as follows:
\begin{align}
\bm r_{\gamma\alpha}&~=~\bm x_{\gamma}-\bm x_{\alpha}, \quad \bm r_{\beta\alpha}~=~\bm x_{\beta}-\bm x_{\alpha},\quad \bm r_{\beta\gamma}~=~\bm x_{\beta}-\bm x_{\gamma}, \nonumber \\  \bm d_{\gamma\alpha}&~=~\bm y_{\gamma}-\bm y_{\alpha}, \quad \bm d_{\beta\alpha}~=~\bm y_{\beta}-\bm y_{\alpha}, \quad \bm d_{\beta\gamma}~=~\bm y_{\beta}-\bm y_{\gamma}.
\end{align}

A linear transformation function ${\bm F}_d$ is assumed for each Delaunay triangle ${\cal T}_{\alpha\beta\gamma}$, that maps the reference and the deformed configurations such that $\bm d_{\beta\alpha} ={\bm F}_d{\bm r_{\beta\alpha}}$ and $\bm d_{\gamma\alpha} ={\bm F}_d{\bm r_{\gamma\alpha}}$.
By solving the pair of equations for each triangle, the discrete deformation tensor ${\bm F}_d$ expressed as a function of particle positions $\bm{x_i}$ and $\bm{y_i} \quad i = \alpha, \beta,~ \text{and}~ \gamma$,
\begin{equation}
{\bm F}_d=\left(\bm d_{\beta\alpha}\otimes\bm e_1+\bm d_{\gamma\alpha}\otimes\bm e_2\right)\left(\bm r_{\beta\alpha}\otimes\bm e_1+\bm r_{\gamma\alpha}\otimes\bm e_2\right)^{-1}.
\label{eq:Tdef}
\end{equation}
The expression $\bm a \otimes \bm b$ denotes the dyadic product of the vectors $\bm a$ and $\bm b$. The energy of each triangle is expressed as a function of the discrete analogue of the Lagrangian strain tensor ($\bm{E_d} = \frac{1}{2}(\bm{F}_d^T \bm{F}_d - \bm{I}$) as 
\begin{equation}\label{Eq.TW}
    W_{\alpha\beta\gamma} = V_{\alpha\beta\gamma} \tilde\psi\left({\bm E}_d\right).
\end{equation}

The interparticle interactions are defined from the macroscopic constitutive response of the material through the total energy of the body $W = \sum W_{\alpha\beta\gamma}$, where the sum is taken over all Delaunay triangles in the domain. The free energy is given by $ \bar\psi = \frac{1}{2} \boldsymbol{E}^T \mathbf{C} \boldsymbol{E}$, where $\bm{C}$ is the constitutive matrix of the material.  The interaction force on particle $i$ with all its neighbors is then
\begin{align}
\bm {f}_{i} &= - {\partial {W}}/{\partial \bm{y}_i}~, \quad \text{where} \quad
W = \sum V_{\alpha\beta\gamma} \tilde\psi\left({\bm E}_d\right).
\end{align}

\subsection{Description of brittle crack}\label{Sec:mod_crack}
Brittle failure is described using the maximum principal stress failure criterion. The stress is evaluated for every triangle in the domain using the corresponding Lagrangian strain $\bm{E}_{d}$ and constitutive relation. The maximum principal stress ($\sigma_1 > \sigma_2$) is compared with the critical value,
\begin{equation}
\sigma_1 \le \sigma_t^U; \quad \sigma_2 \ge -\sigma_c^U
\end{equation}
where \( \sigma^t_U ~\text{and}~ \sigma^c_U \) are the ultimate tensile and compressive strengths, respectively. The maximum principal stress ($\sigma_1$) of each triangle is continuously monitored during simulation, and if the triangle stress satisfies the failure criteria $\sigma_1 \ge \sigma_t^U ~\text{or}~ \sigma_2 \le -\sigma_c^U$, the interaction forces associated with that triangle are set to zero, and the corresponding particles no longer interact through that particular triangle.

\begin{figure}[h]
\begin{center}
\includegraphics[scale=0.75]{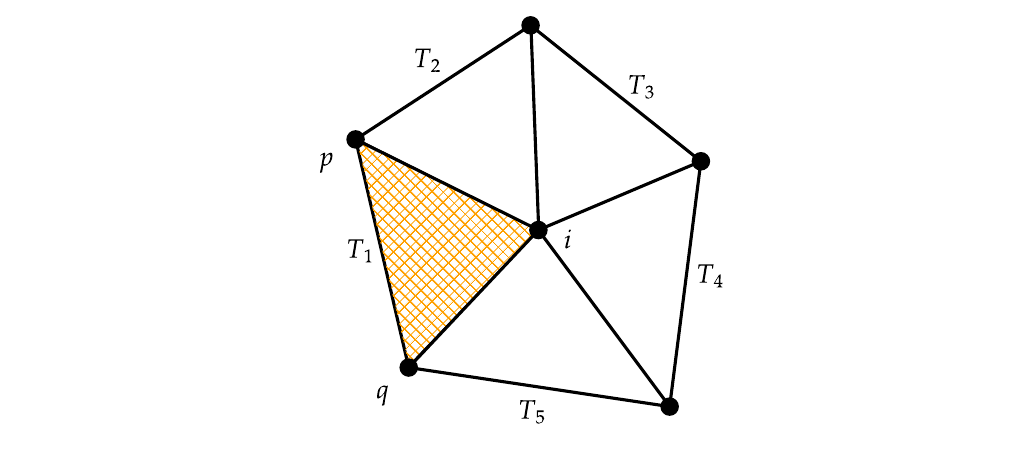}
\caption{The particle $i$ and its neighbors interacting through Delauney triangles $T_i;~ i=1,2 \dots 5$. The strain energy density of $T_1$ at a particular instance shown satisfies the failure criteria $\sigma_1^{T_1} = \sigma^t_U$.}
\label{fig:Crack_model}
\end{center}
\end{figure}
Consider six particles as shown in Figure~\ref{fig:Crack_model} which are interacting through the respective Delauney triangles. Since particle $i$ is interacting with all other five neighboring particles through triangles $T_1  - T_5$ the interaction force on particle $i$ can be obtained by 
\begin{equation}
{\bm f_i} = \sum_{j=1}^5 - \frac{\partial {W_{T_j}}}{\partial \bm{y}_i}; \quad W_{T_j} =  W_{\alpha \beta \gamma},
\end{equation}
$W_{T_j}$ denote the strain energy of the triangle $T_j$ and it is evaluated using $W_{\alpha \beta \gamma}$ from Eq. \ref{Eq.TW} where $\alpha, \beta,$ and $\gamma$ denote the corresponding particles in the respective triangle ${T_j}$. Specifically the particles $p$, $q$ and $i$ are interacting through triangle $T_1$. While the system evolves, consider the failure condition $\sigma_1^{T_1} = \sigma^t_U$ is satisfied for a triangle $T_1$. Then the interaction force from triangle $T_1$ on particles $p$, $q$ and $i$ are made zero. The particle interaction between $p$ and $q$ are completely removed but the particles $p$ and $i$ interacting through $T_2$ and the particles $q$ and $i$ interacting through $T_5$.

\section{Modeling geometric discontinuity and brittle crack}\label{sec:Modeling geometric discontinuity}

\begin{figure}[!tbh]
    \centering
    \begin{minipage}{0.3\textwidth}
        \centering
        {\includegraphics[width=0.9\textwidth, trim= -0.9cm 0 0 0, clip]{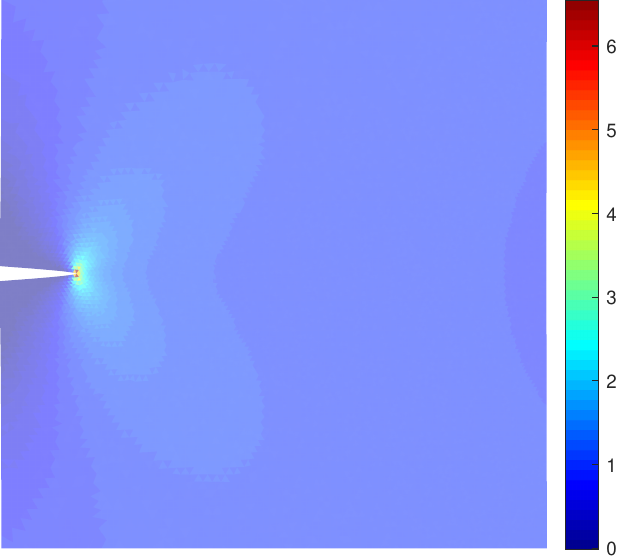}}
        {\includegraphics[width=1\textwidth, trim= 0 0 -1.2cm 0, clip]{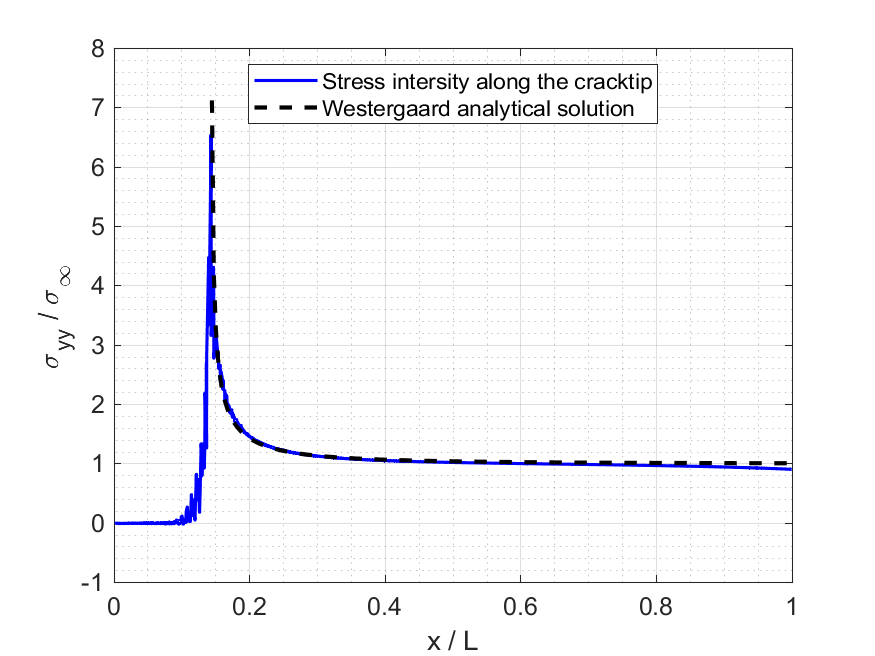}}
        \subcaption{Stress concentration near the crack tip.}\label{fig:validation_stress_concentration_crack}
    \end{minipage}\hfill
    \centering
    \begin{minipage}{0.3\textwidth}
        \centering
        {\includegraphics[width=0.9\textwidth, trim= -9mm 0 0 0, clip]{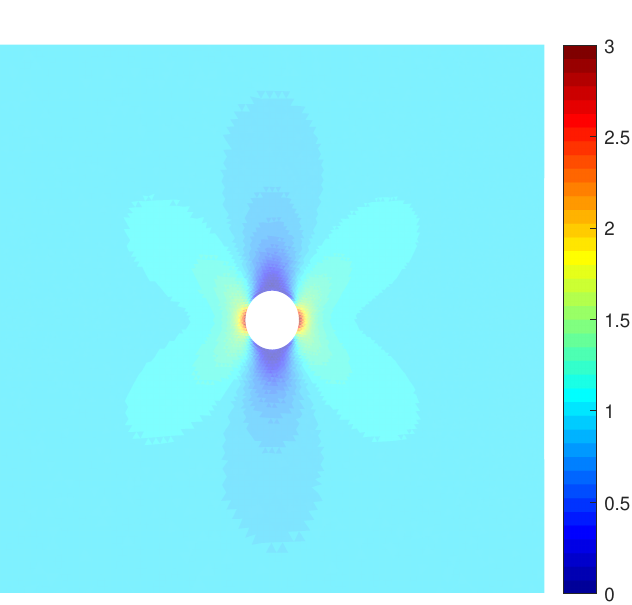}}
        {\includegraphics[width=1\textwidth, trim= 0 0 -12mm 0, clip]{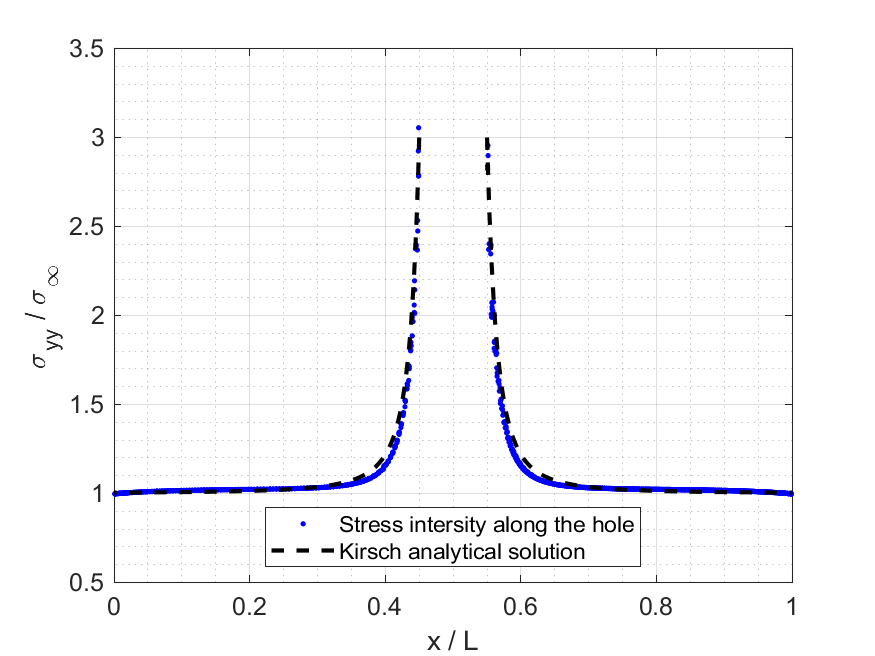}}
        \subcaption{Stress concentration around the hole.}\label{fig:validation_stress_concentration_hole}
    \end{minipage}\hfill
    \centering
    \begin{minipage}{0.3\textwidth}
        \centering
        {\includegraphics[width=0.9\textwidth, trim= -9mm 0 0 0, clip]{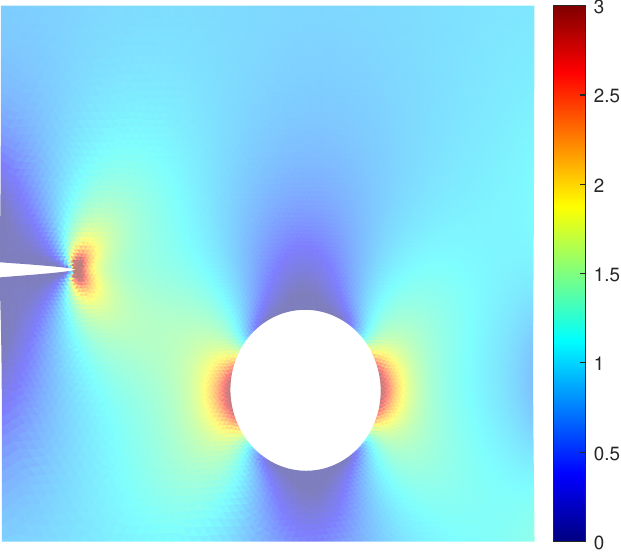}}
        {\includegraphics[width=1\textwidth, trim= 0 0 -12mm 0, clip]{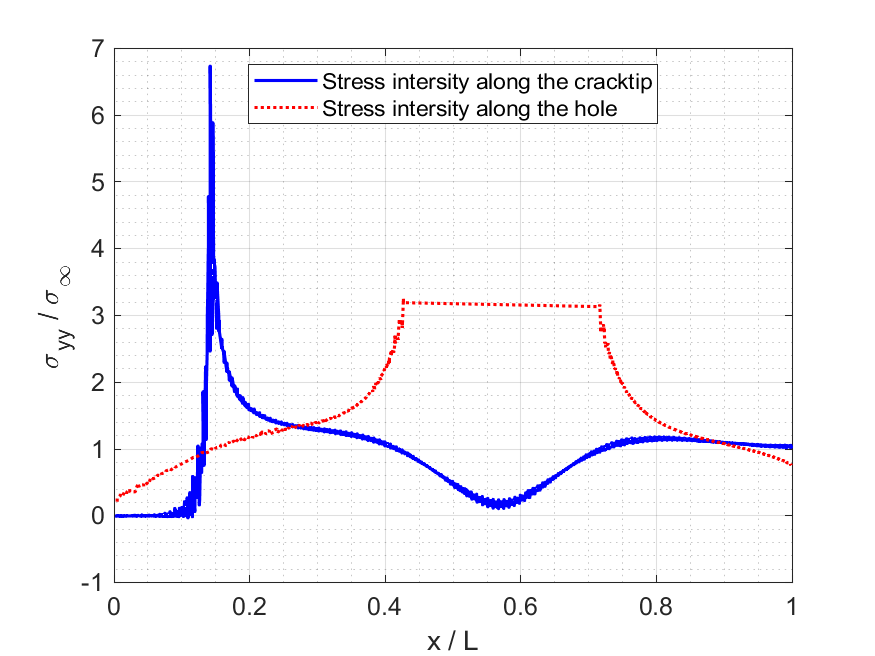}}
        \subcaption{Stress concentration near the cracktip and hole.}\label{fig:validation_stress_concentration_interaction}
    \end{minipage}\hfill
    \caption{Stress field concentrations around geometric discontinuities in the domain.}
    \label{fig:validation_stress_concentration}
\end{figure}

To demonstrate the model’s ability to capture stress concentrations around discontinuities, three benchmark cases were considered: a pre-existing sharp crack (Figure \ref{fig:validation_stress_concentration_crack}), a circular hole (Figure \ref{fig:validation_stress_concentration_hole}), and a circular hole interacting with a pre-existing crack (Figure \ref{fig:validation_stress_concentration_interaction}). The domain is discretized with approximately 16,000 particles with over 30,000 multi-body interaction triangulations that are identified using Delaunay triangulation as shown in Figure \ref{fig:geometry:mesh}. Displacement-based loading is applied to particles on the top and bottom boundaries, incremented in small steps until a predefined value is reached to ensure a quasi-static loading scenario. Newton’s equations of motion are solved for each particle using an isotropic elastic strain energy function, with Young’s modulus \(E = 210\,\text{GPa}\) and Poisson’s ratio \(\nu = 0.3\), as presented in \cite{CPD_CMAME}.

The distribution of stress concentration around the crack tip and circular hole is shown in Figure \ref{fig:validation_stress_concentration_crack}, and Figure~\ref{fig:validation_stress_concentration_hole}, respectively. The stress profiles along the crack tip and the horizontal line passing through the center of the circular hole are compared with analytical solutions given by Westergaard \cite{westergaard1939bearing} and Kirsch \cite{kirsch1898theorie}, respectively. The numerical results exhibit good agreement with the analytical expressions, given in Eq.~\ref{eq:Westergaard} for the crack tip and Eq.~\ref{eq:Kirsch} for the hole. Figure \ref{fig:validation_stress_concentration_interaction} further illustrates the stress distribution and interaction in the domain containing both a crack and a circular hole.

\begin{equation}
   \sigma_{yy} = \frac{\sigma_\infty}{\sqrt{1 - \left( \frac{a}{x} \right)^2}} 
\label{eq:Westergaard} 
\end{equation}

\begin{equation}    
    \sigma_{yy} =  \frac{\sigma_\infty}{2} \left( 1 + \left( \frac{r}{x} \right)^2 \right) + \frac{\sigma_\infty}{2} \left( 1 + 3 \left( \frac{r}{x} \right)^4 \right) 
\label{eq:Kirsch} 
\end{equation}

where $a$ is the length of the pre-existing crack, $r$ is the radius of the hole, and $x$ is the horizontal distance along the discontinuity, as shown in Fig.~\ref{fig:geometry:domain}.

\begin{figure}[!tbh]
    \centering
    \begin{minipage}{0.3\textwidth}
        \centering
        \includegraphics[width=0.85\textwidth, trim=1cm 0.75cm 1.5cm 0.5cm, clip]{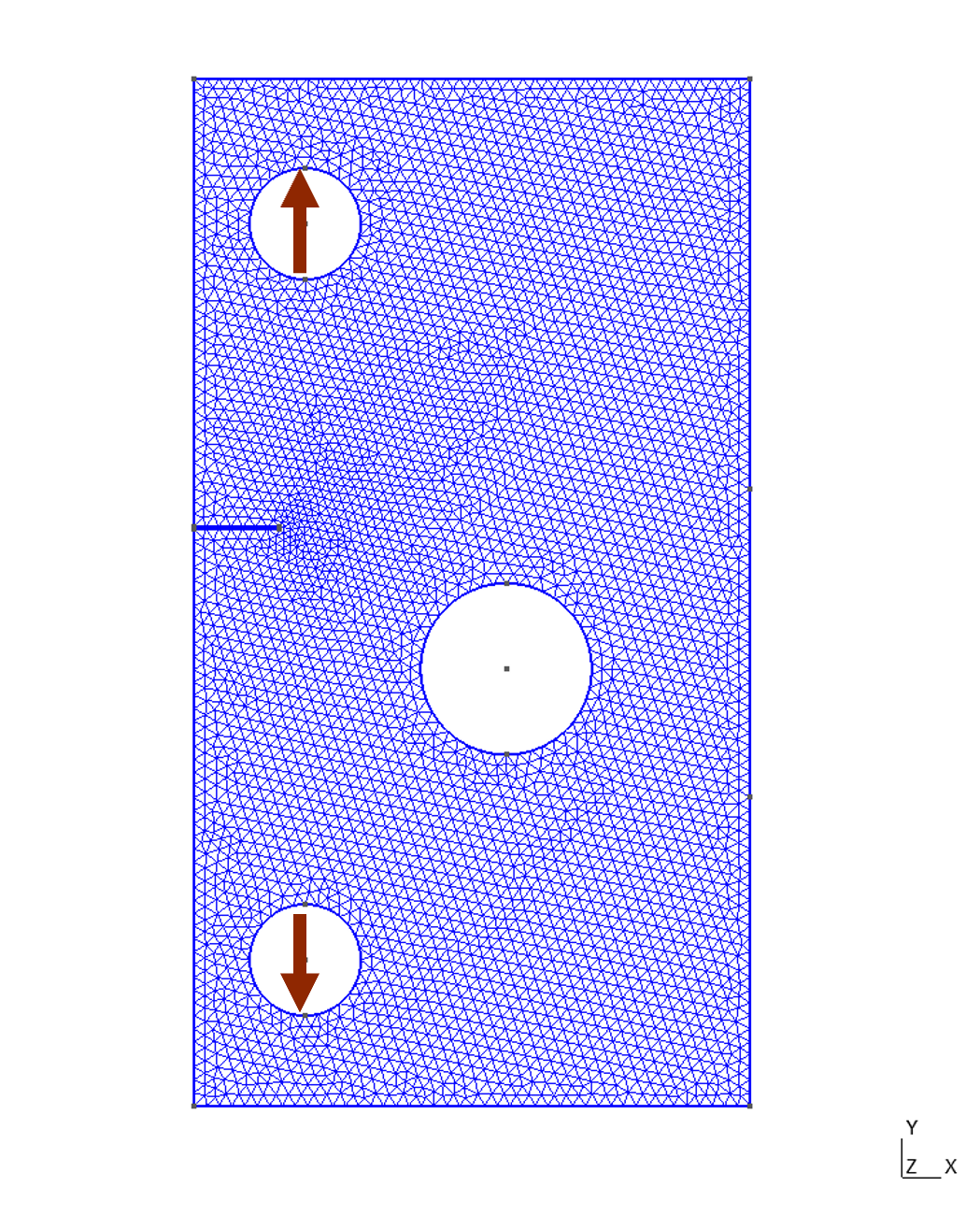}
        \subcaption{Domain geometry and boundary conditions used in the simulation.}
        \label{fig:validation_exp_mesh}
    \end{minipage}\hfill
    \begin{minipage}{0.3\textwidth}
        \centering
        \includegraphics[width=0.8\textwidth]{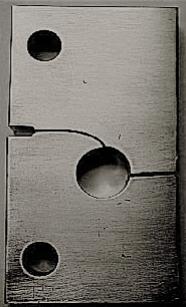}
        \subcaption{Experimentally observed crack path reproduced from \cite{2009abaqus}.}
        \label{fig:validation_exp_ref}
    \end{minipage}\hfill
    \begin{minipage}{0.3\textwidth}
        \centering
        \includegraphics[width=0.8\textwidth, trim=7.5cm 7.5cm 6.5cm 7cm, clip]{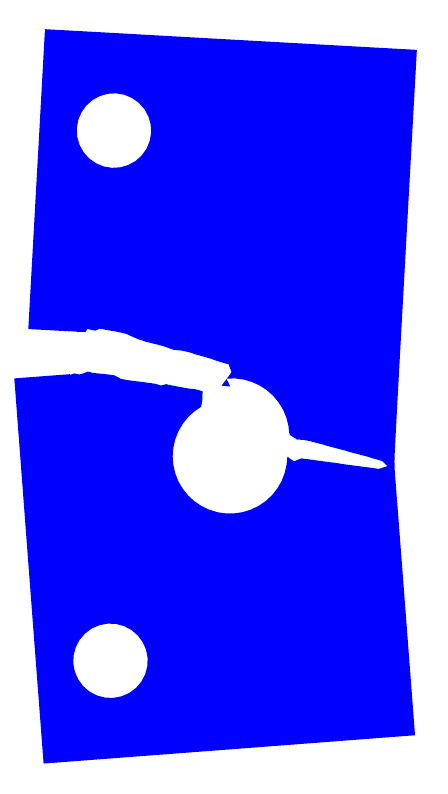}
        \subcaption{Simulated deformation and crack path prior to complete separation.}
        \label{fig:validation_exp_results}
    \end{minipage}
    \caption{Comparison of predicted crack path with experimental results and domain setup.}
    \label{fig:validation_exp}
\end{figure}

Similarly, to demonstrate the model’s capability in capturing crack initiation and propagation due to the interaction of stress fields between a pre-existing crack tip and a nearby hole, the fracture mechanism is incorporated as described in Section~\ref{Sec:mod_crack}. Figure~\ref{fig:validation_exp} illustrates the model’s ability to predict the crack path, which aligns well with experimental observations. Figure~\ref{fig:validation_exp_mesh} shows the domain geometry and applied boundary conditions. The experimentally observed crack trajectory is shown in Figure~\ref{fig:validation_exp_ref}, reproduced from \cite{2009abaqus}. The numerically predicted deformation and crack propagation path, just before complete separation, are presented in Figure~\ref{fig:validation_exp_results}, demonstrating excellent agreement with the experimental reference.

\subsection{Influence of geometric discontinuities on crack path}

\begin{figure}[!tbh]
    \centering
    \begin{minipage}{0.35\textwidth}
        \centering
        {\includegraphics[width=0.7\textwidth, trim= 0 0 5cm 0, clip]{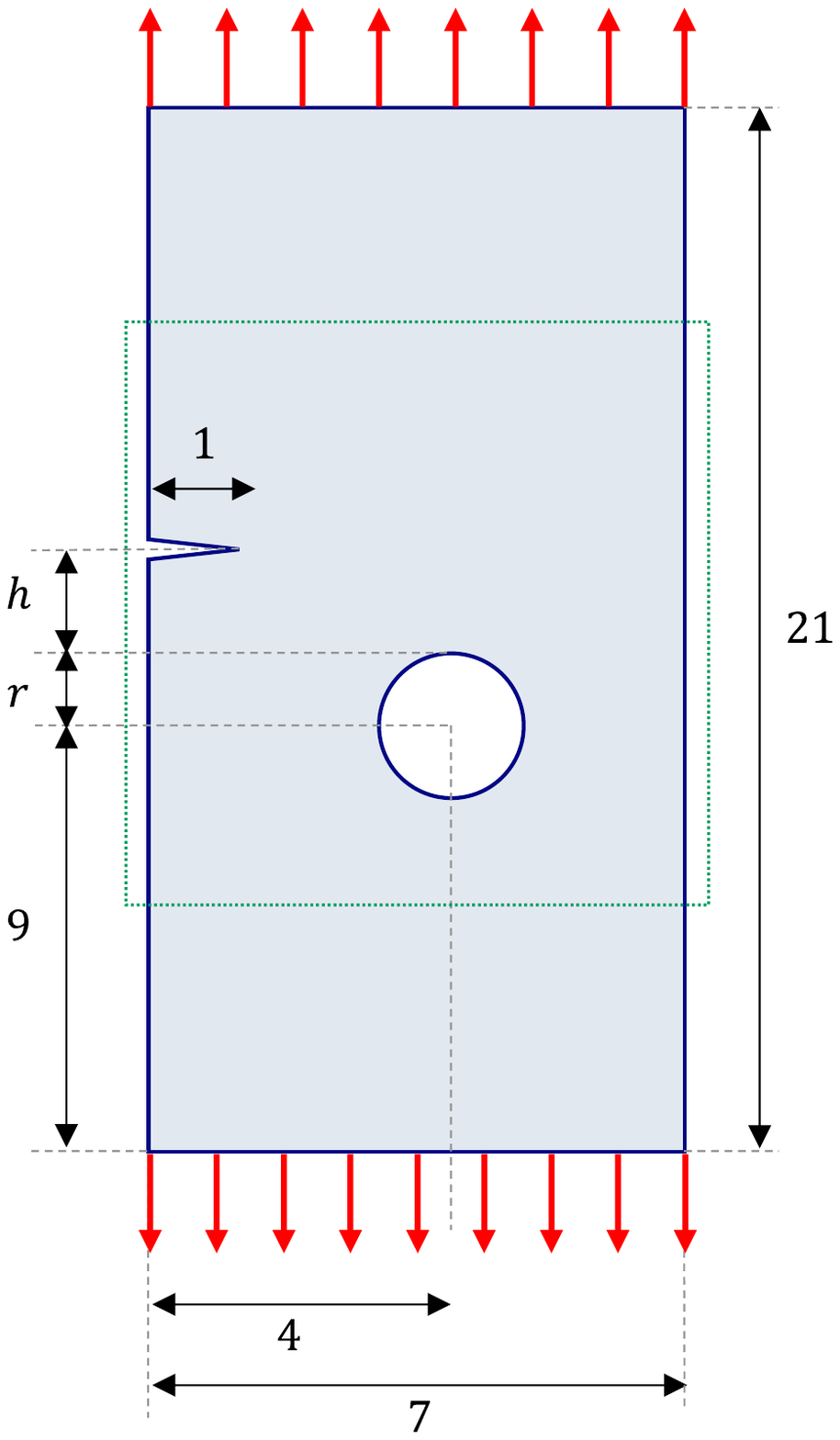}}
        \subcaption{ }\label{fig:geometry:domain}
    \end{minipage}\hfill
    \centering
    \begin{minipage}{0.65\textwidth}
        \centering
        {\includegraphics[width=0.6\textwidth, trim= 0 0 0 0, clip]{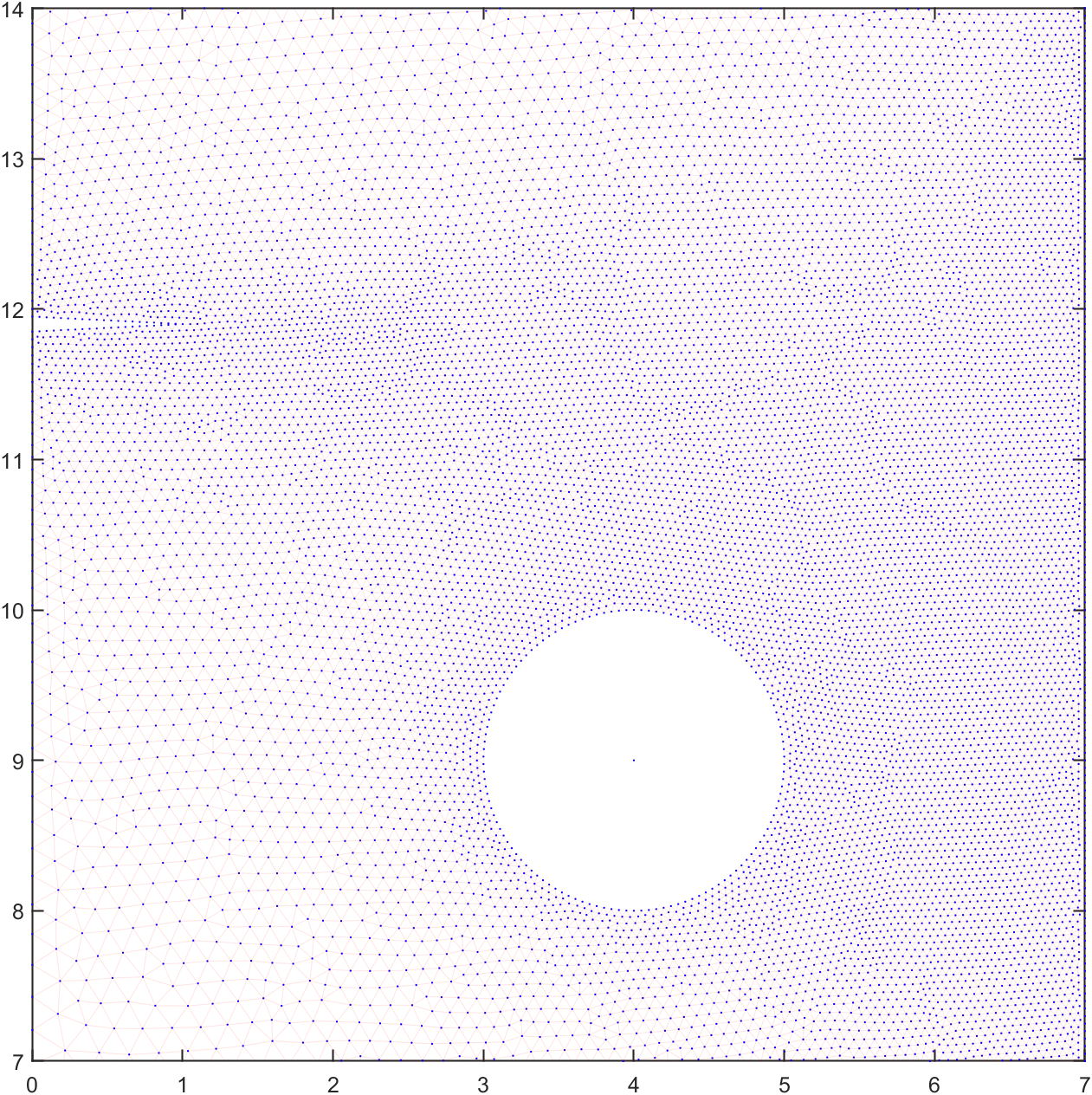}}
        \subcaption{} \label{fig:geometry:mesh}
    \end{minipage}\hfill
    \caption{(a) Geometry of domain with hole and pre-existing crack along with the loading conditions. The green dotted box highlights the region for crack path analysis. (b) Reference particle configuration and multi-body interaction triangulation for the zoomed-in region.}

    \label{fig:geometry}
\end{figure}

The rectangular domain with a pre-existing crack and a circular hole, adopted from \cite{rubinstein1991mechanics}, is considered for the analysis, as shown in Figure~\ref{fig:geometry}. The green box in Figure~\ref{fig:geometry:domain} indicates the region of interest where crack propagation is examined, and subsequent figures show a cropped view of this area. The domain is discretized using approximately 16,000 particles, forming over 30,000 multi-body interaction triangulations, as illustrated in Figure~\ref{fig:geometry:mesh}.

To simulate and generate data for training and testing the operator learning framework on particle displacement and crack evolution, three representative cases are considered. These cases vary in geometric parameters and fracture assumptions, as summarized in Table~\ref{tab:case_Cases}. For each case, the reference particle configuration is recorded at $ \tau = 0 $, and the deformed configurations are recorded over 100 equally spaced simulation stages, from $ \tau = 1 $ to $ \tau = 100 $, for use in training and testing the model. Here, $ \tau \in [0, 100] $ denotes the deformation time-step index, representing 100 equally spaced steps from the undeformed configuration ($ \tau = 0 $) to the fully deformed state ($ \tau = 100 $).

\begin{table}[ht]
\renewcommand{\arraystretch}{1.01} 
\centering
\caption{Geometric and mesh details for three representative case studies.}
\begin{tabular}{|>{\centering\arraybackslash}p{2.1cm}|
                >{\centering\arraybackslash}p{4.2cm}|
                >{\centering\arraybackslash}p{2.7cm}|
                >{\centering\arraybackslash}p{2.7cm}|
                >{\centering\arraybackslash}p{2.6cm}|
                >{\centering\arraybackslash}p{2.6cm}|}
\hline
\textbf{Case} & \textbf{Parametrization Strategy} & \textbf{Radius (cm)} & \textbf{Height (cm)} & \textbf{Number of Samples} & \textbf{Fracture Accounted} \\
\hline
\textbf{Case 1}  & Varying pre-crack notch height & $r = 1$ & $h = 0$ to $2$ & 40 & No \\
\hline
\textbf{Case 2} & Varying pre-crack notch height & $r = 1$ & $h = 0$ to $2$ & 50 & Yes \\
\hline
\textbf{Case 3} & Varying hole radius & $r = 0.5$ to $1.5$ & $h = 1.5$ & 51 & Yes \\
\hline
\end{tabular}
\label{tab:case_Cases}
\end{table}

 In \textbf{Case 1}, a total of forty rectangular specimens are analyzed, each containing a circular hole of fixed radius \( r = 1~\text{cm} \). The height parameter \( h \), which defines the vertical distance between the pre-existing horizontal crack and the circular hole, is systematically varied from \( h = 0~\text{cm} \) to \( h = 2~\text{cm} \). These simulations are carried out without assuming any failure criteria, thereby focusing purely on elastic deformation and particle displacement without fracture. In \textbf{Case 2}, the same rectangular domain geometry is retained with a fixed circular hole of radius \( r = 1~\text{cm} \), but this time fifty different configurations are considered by varying the height \( h \) from \( h = 0~\text{cm} \) to \( h = 2~\text{cm} \). Unlike the first case, these simulations incorporate fracture evolution, enabling the study of crack initiation and propagation in relation to the geometry. \textbf{Case 3} explores the influence of the hole radius on crack behavior. Here, the vertical distance \( h \) is fixed at \( h = 1.5~\text{cm} \), and the radius \( r \) of the circular hole is varied from \( r = 0.5~\text{cm} \) to \( r = 1.5~\text{cm} \). Fifty-one unique specimens are simulated in this case, all of which include fracture modeling.

\begin{figure}[!tbh]
    \centering
    \begin{minipage}{0.3\textwidth}
        \centering
        \includegraphics[width=\textwidth, trim=4cm 7cm 4cm 7cm, clip]{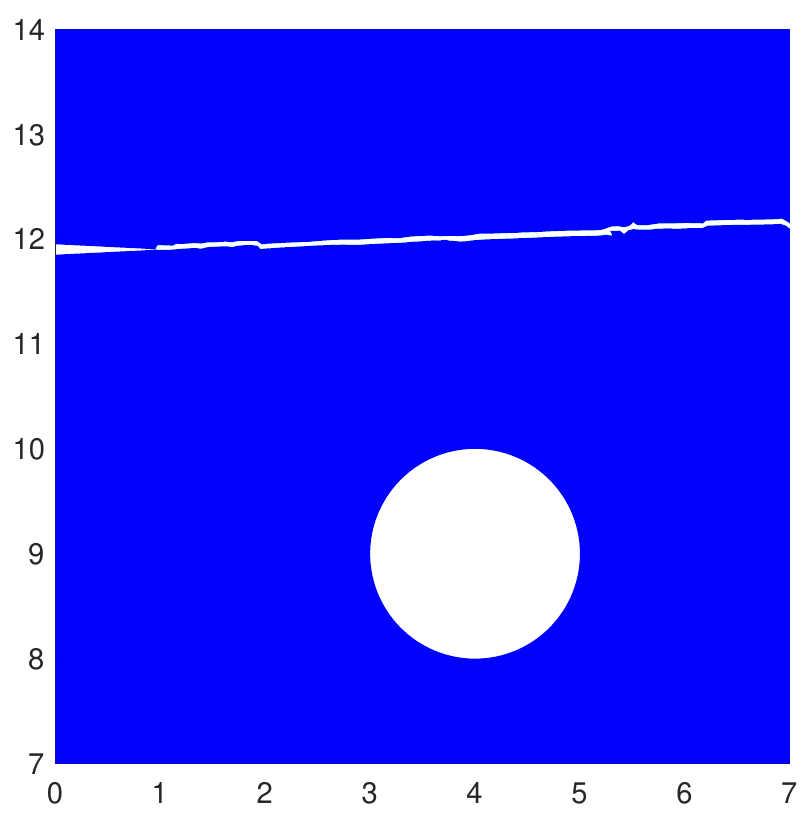}
        \subcaption{$h = 1.90~cm$}\label{fig:hhole:a}
    \end{minipage}\hfill
    \begin{minipage}{0.3\textwidth}
        \centering
        \includegraphics[width=\textwidth, trim=4cm 7cm 4cm 7cm, clip]{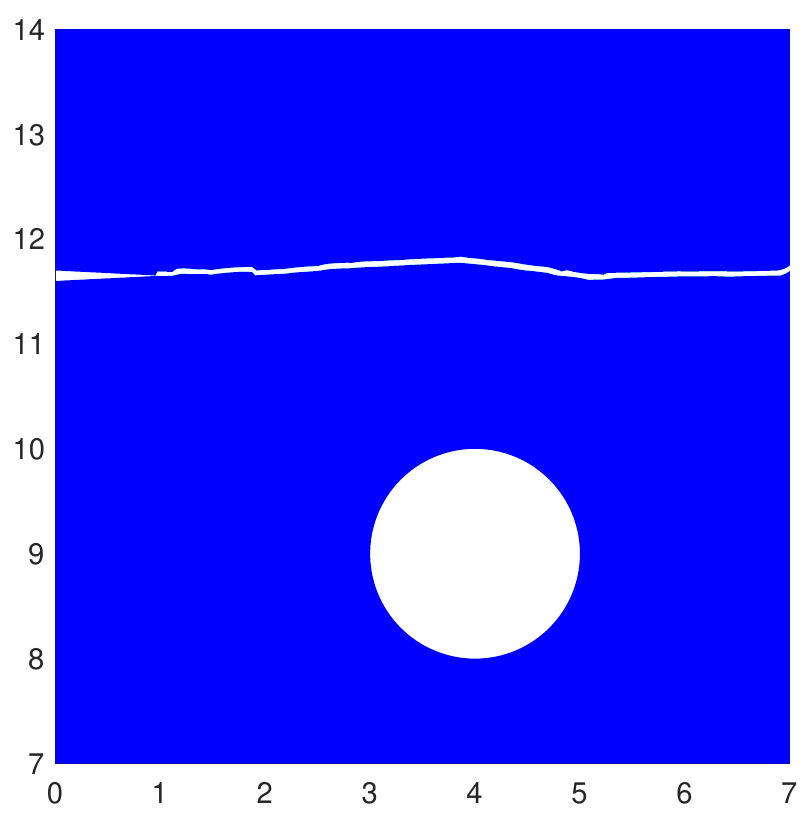}
        \subcaption{$h = 1.65~cm$}\label{fig:hhole:b}
    \end{minipage}\hfill
    \begin{minipage}{0.3\textwidth}
        \centering
        \includegraphics[width=\textwidth, trim=4cm 7cm 4cm 7cm, clip]{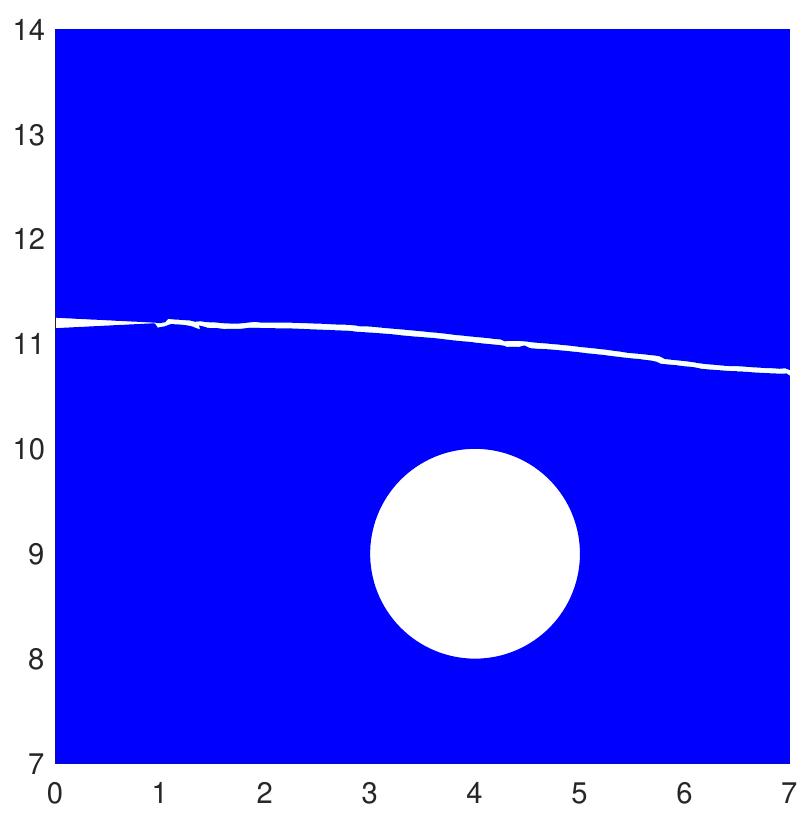}
        \subcaption{$h = 1.20~cm$}\label{fig:hhole:c}
    \end{minipage}
    
    \vskip 1em 

    \begin{minipage}{0.3\textwidth}
        \centering
        \includegraphics[width=\textwidth, trim=4cm 7cm 4cm 7cm, clip]{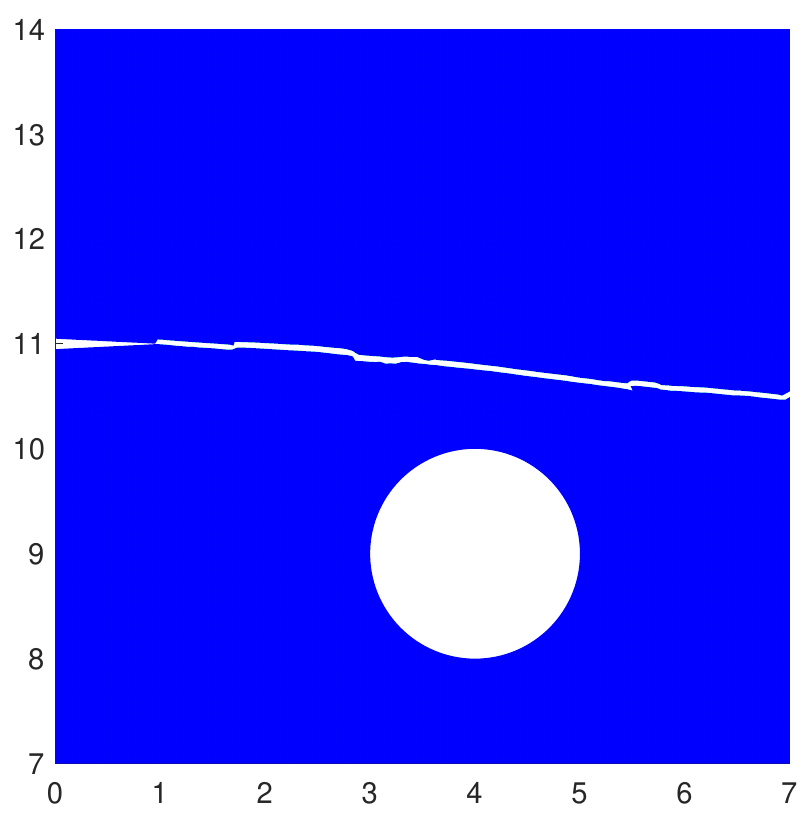}
        \subcaption{$h = 1.00~cm$}\label{fig:hhole:d}
    \end{minipage}\hfill
    \begin{minipage}{0.3\textwidth}
        \centering
        \includegraphics[width=\textwidth, trim=4cm 7cm 4cm 7cm, clip]{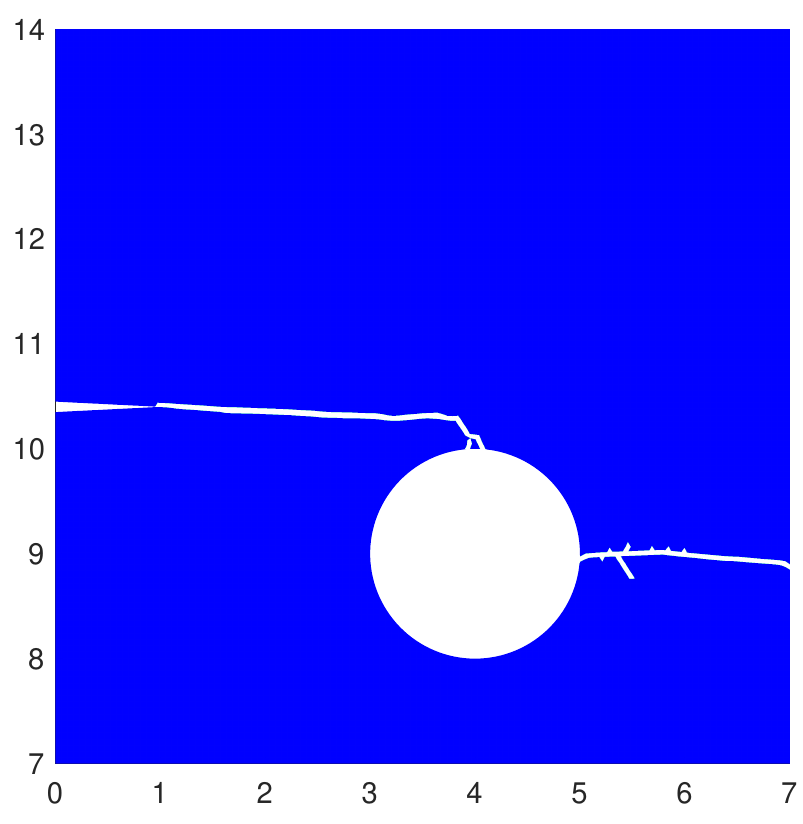}
        \subcaption{$h = 0.40~cm$}\label{fig:hhole:e}
    \end{minipage}\hfill
    \begin{minipage}{0.3\textwidth}
        \centering
        \includegraphics[width=\textwidth, trim=4cm 7cm 4cm 7cm, clip]{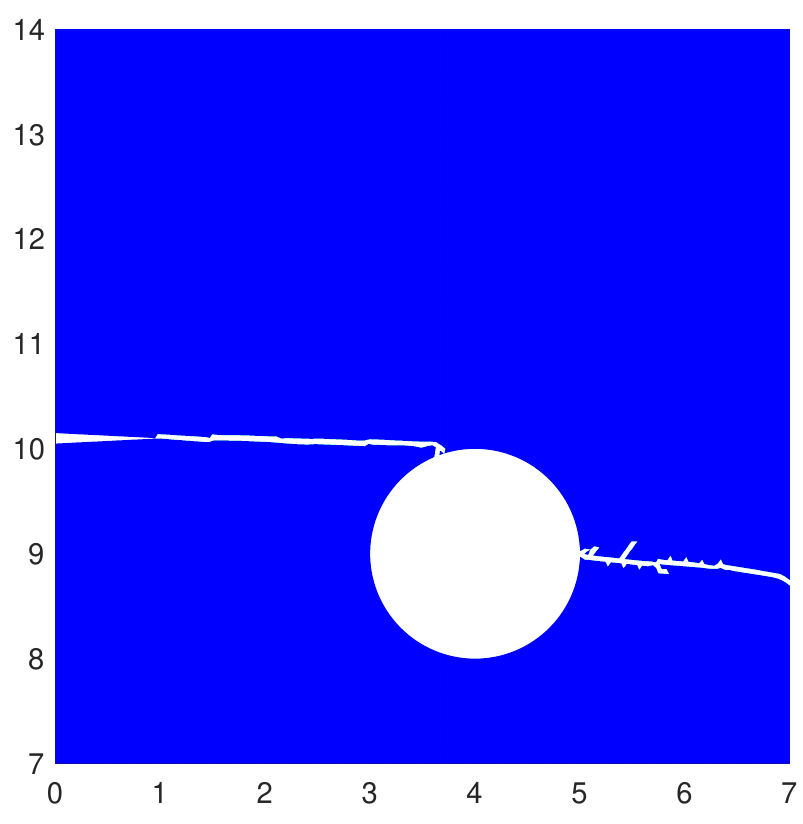}
        \subcaption{$h = 0.10~cm$}\label{fig:hhole:f}
    \end{minipage}

    \caption{\textbf{Case 2: } Crack paths simulated by varying the height ($h$) of pre-existing crack from the hole of constant radius of $r= 1~cm$.}
    \label{fig:Case2}
\end{figure}

\begin{figure}[!tbh]
    \centering
    \begin{minipage}{0.3\textwidth}
        \centering
        \includegraphics[width=\textwidth, trim=4cm 7cm 4cm 7cm, clip]{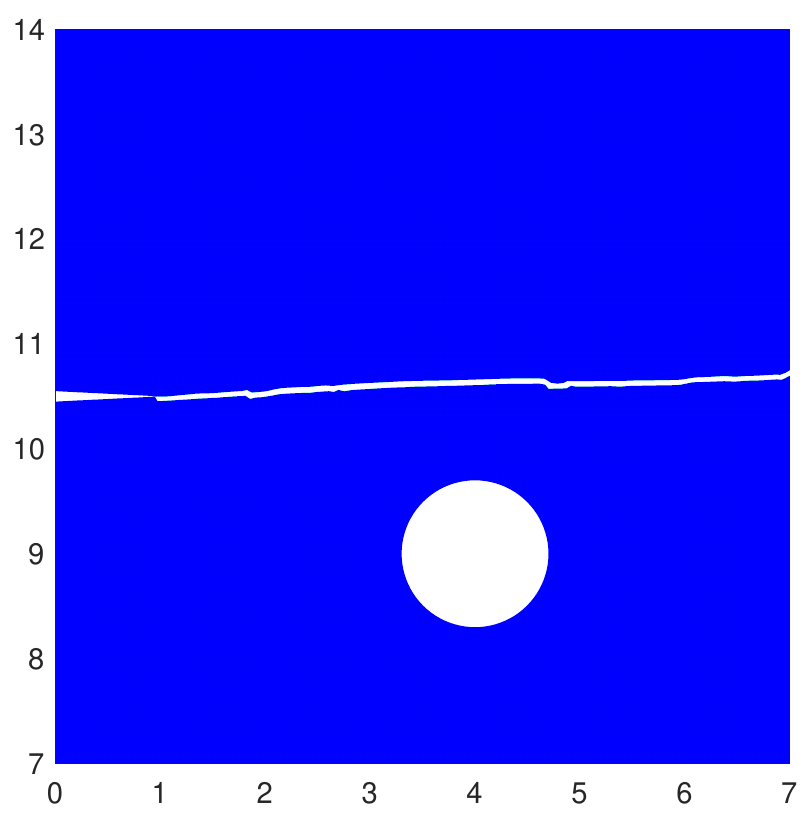}
        \subcaption{$r = 0.70 ~cm$}\label{fig:rrhole:a}
    \end{minipage}\hfill
    \begin{minipage}{0.3\textwidth}
        \centering
        \includegraphics[width=\textwidth, trim=4cm 7cm 4cm 7cm, clip]{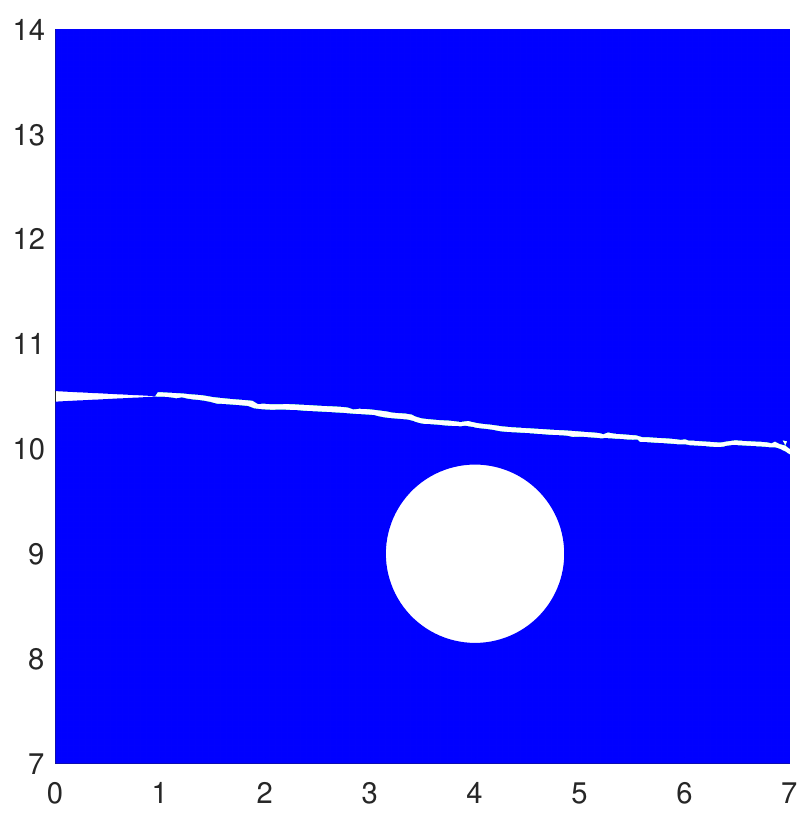}
        \subcaption{$r = 0.85 ~cm$}\label{fig:rrhole:b}
    \end{minipage}\hfill
    \begin{minipage}{0.3\textwidth}
        \centering
        \includegraphics[width=\textwidth, trim=4cm 7cm 4cm 7cm, clip]{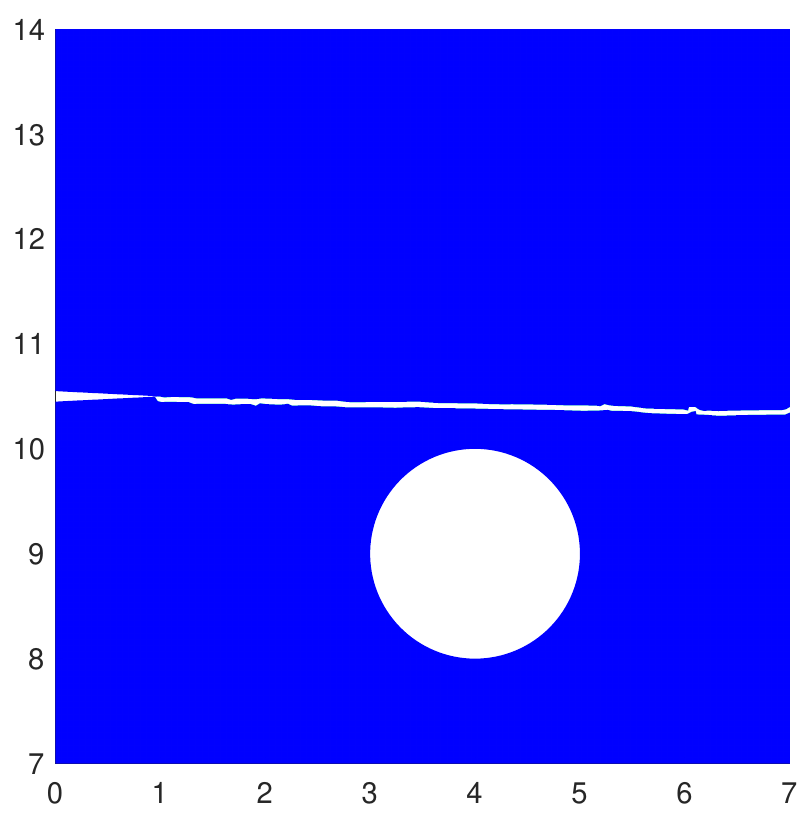}
        \subcaption{$r = 1.00 ~cm$}\label{fig:rrhole:c}
    \end{minipage}
    
    \vskip 1em 

    \begin{minipage}{0.3\textwidth}
        \centering
        \includegraphics[width=\textwidth, trim=4cm 7cm 4cm 7cm, clip]{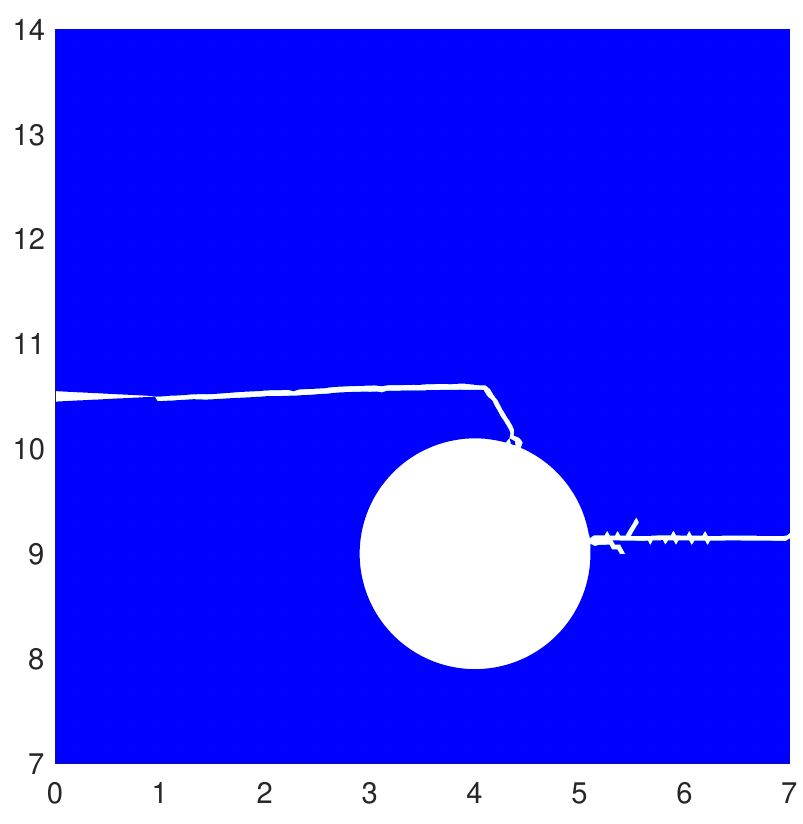}
        \subcaption{$r = 1.1~cm$}\label{fig:rrhole:d}
    \end{minipage}\hfill
    \begin{minipage}{0.3\textwidth}
        \centering
        \includegraphics[width=\textwidth, trim=4cm 7cm 4cm 7cm, clip]{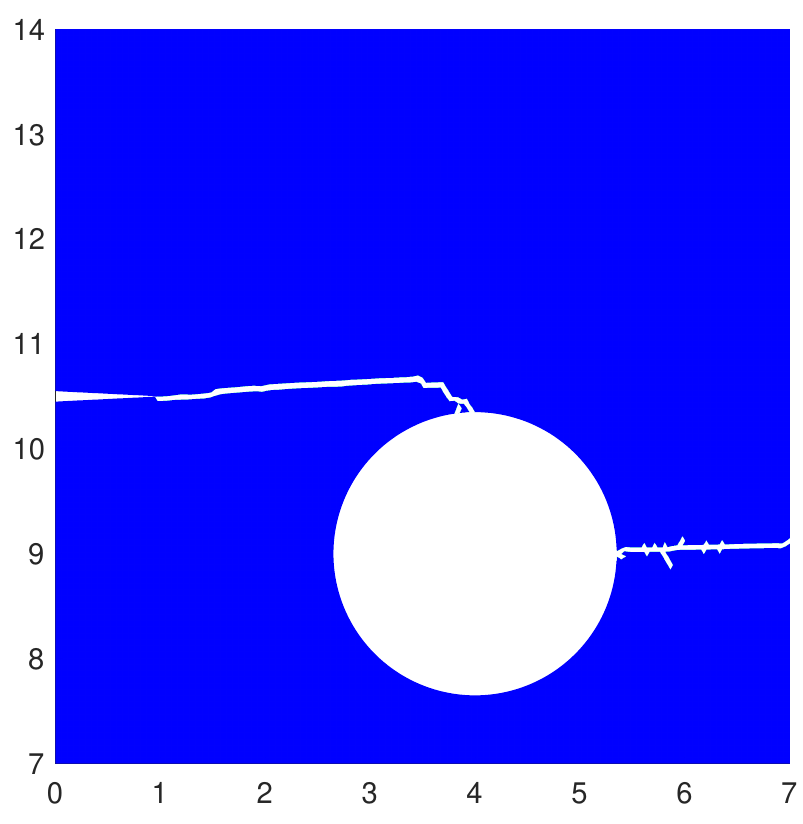}
        \subcaption{$r = 1.35~cm$}\label{fig:rrhole:e}
    \end{minipage}\hfill
    \begin{minipage}{0.3\textwidth}
        \centering
        \includegraphics[width=\textwidth, trim=4cm 7cm 4cm 7cm, clip]{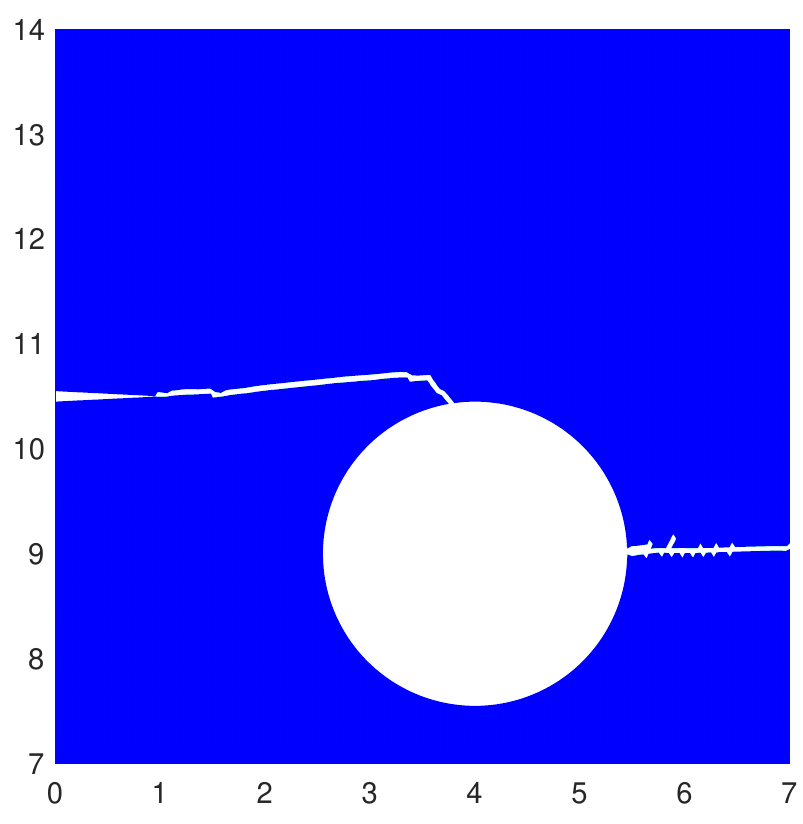}
        \subcaption{$r = 1.45~cm$}\label{fig:rrhole:f}
    \end{minipage}

    \caption{\textbf{Case 3:} Crack paths simulated by varying the radius ($r$) of the hole positioned at $h = 1.5~cm$ below the pre-existing crack.}

    \label{fig:case3}
\end{figure}

Simulated crack paths for selected specimens from the second and third cases are shown in Figures~\ref{fig:Case2}(a–f) and \ref{fig:case3}(a–f), respectively, focusing on the zoomed-in region highlighted by the green dotted box in Figure~\ref{fig:geometry}. In this analysis, the key geometric parameters influencing crack propagation include the initial crack length \( a \), the radius of the circular hole \( r \), and the vertical separation \( h \) between the crack tip and the hole center. For all simulations, the crack length is fixed at \( a = 1~\text{cm} \). It is observed that as the vertical distance \( h \) is reduced, the crack begins to deviate from its original direction, increasingly interacting with the stress field generated by the hole. This interaction can result in the crack deflecting toward, and eventually merging with, the hole. A similar effect is noted as the radius \( r \) increases, it promotes crack deflection and eventual coalescence. For a fixed pre-existing crack length \( a = 1~\text{cm} \), this transition is consistently observed when the dimensionless parameter \( (r + h)/r \leq 1.4 \), where the geometric configuration favors crack–hole interaction, often resulting in the crack being trapped by the hole and eventual passing through the hole on further loading.

\section{Neural Operator Learning with DeepONet}
\label{sec:Operator_Learning}

In conventional machine learning, models are typically trained to learn relationships between finite-dimensional vectors. However, such formulations are often inadequate for scientific and engineering problems that require learning mappings between functions defined over continuous domains—that is, in infinite-dimensional spaces. Operator learning overcomes this limitation by approximating mappings of the form
\[
\mathcal{G} : u(\xi) \mapsto v(\xi),
\]
where both the input \( u(\xi) \) and the output \( v(\xi) \) are functions defined on a continuous domain~\cite{lu2019deeponet,peyvan2025fusion,kiyani2024predicting}.
This framework is particularly well-suited for surrogate modeling of parametric partial differential equations (PDEs), where the objective is to learn a solution operator that maps input conditions—such as initial states, boundary values, or source terms—to the corresponding output fields.

\subsection{DeepONet Setup}

Let \( \mathcal{G} : \mathcal{A} \rightarrow \mathcal{B} \) be a (possibly nonlinear) operator between function spaces \( \mathcal{A} \) and \( \mathcal{B} \), defined over a spatial domain \( D \subset \mathbb{R}^d \). Given a set of training examples \( \{(a_i, b_i)\}_{i=1}^N \), where \( b_i = \mathcal{G}(a_i) \), the goal is to learn an approximation \( \hat{\mathcal{G}} \) such that
\[
\hat{\mathcal{G}}(a) \approx \mathcal{G}(a), \quad \text{for all } a \in \mathcal{A}.
\]
In other words, the learned operator should generalize well to both training and unseen input functions.

DeepONet is a neural network architecture designed to approximate such operators. It consists of two subnetworks:
\begin{itemize}
    \item \textbf{Branch network:} This network encodes the input function \( f \), evaluated at a fixed set of sensor points \( \{\xi_1, \xi_2, \ldots, \xi_k\} \), and maps it to a latent feature vector \( \bm{b} \in \mathbb{R}^p \). For multiple input samples, this yields a matrix \( B \in \mathbb{R}^{m \times p} \), where each row corresponds to one input function.
    
    \item \textbf{Trunk network:} This network takes an evaluation point \( \xi \in D \) and produces a coordinate-dependent feature vector \( \bm{\phi}(\xi) \in \mathbb{R}^p \), forming a learned basis for reconstructing the output. For a set of evaluation points, it generates a matrix \( \Phi \in \mathbb{R}^{n \times p} \).
\end{itemize}

The final prediction of the operator applied to input function \( f_i \) at a point \( \xi_j \) is given by the inner product of the branch and trunk outputs:
\begin{equation}
\mathcal{G}(f_i)(\xi_j) \approx \sum_{l=1}^{p} B_{il}(f_i; \theta_B) \cdot \Phi_{jl}(\xi_j; \theta_T),
\label{eq:deeponet_outputs}
\end{equation}
where \( \theta_B \) and \( \theta_T \) are the trainable parameters of the branch and trunk networks, respectively.

The displacement field \( \bm{u}(\bm{x}_i, t) \) is defined as the difference between the deformed configuration \( \bm{y}_i(t) \) and the reference configuration \( \bm{x}_i \) of particle \( i \). Mathematically, this is expressed as
\[
\bm{u}(\bm{x}_i, t) = \bm{y}_i(t) - \bm{x}_i,
\]
where \( \bm{x}_i = \bm{y}_i(0) \) denotes the initial (undeformed) position. This formulation is implemented numerically by computing the displacement at each time step as the difference between the current and initial particle positions.

\begin{figure}[!tbh]
    \centering
    \includegraphics[width=0.99\textwidth]{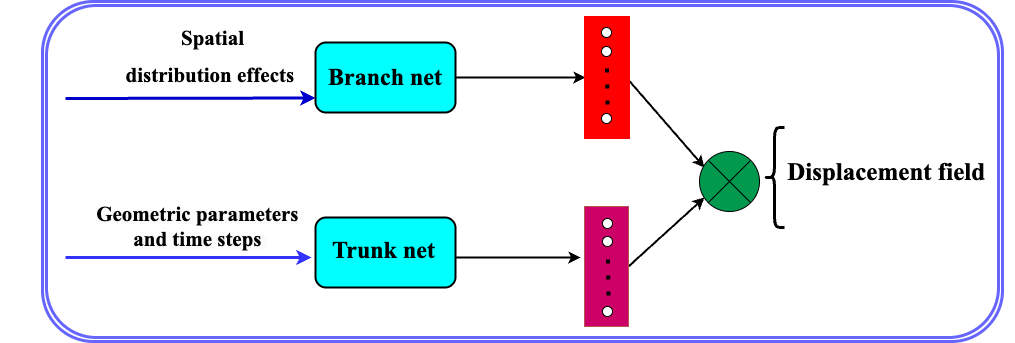}
\caption{Schematic of the DeepONet architecture for displacement field prediction. The branch network encodes input functions related to spatial variations, such as pre-crack notch height or hole radius. The trunk network processes the spatial coordinate \( \bm{x} \in \mathbb{R}^2 \) and time \( t \). The output is the predicted displacement field \( \bm{u}(\bm{x}, t) \).}
    \label{fig:Deeponet_Schematic}
\end{figure}

Figure~\ref{fig:Deeponet_Schematic} illustrates the DeepONet architecture used in this work. The decomposition in~\eqref{eq:deeponet_outputs} decouples the encoding of the input function from the spatial-temporal representation of the output, making DeepONet a powerful and flexible framework for learning nonlinear operators in complex physical systems. 
In this setup, the input to the branch network consists of the initial geometry parameters, such as the pre-crack notch height \( h_i \) and the hole radius \( r_i \). The trunk network takes as input the spatial coordinate \( \bm{x} \in \mathbb{R}^2 \) and time \( t \). The final output is the predicted displacement field \( \bm{u}(\bm{x}, t) \).

\subsection{Fusion-DeepONet}

Fusion DeepONet is an advanced variant of the standard DeepONet, designed to improve data efficiency and predictive accuracy in geometry-dependent problems defined on arbitrary or irregular grids~\cite{peyvan2025fusion}. Like the original architecture, Fusion DeepONet consists of two subnetworks: a branch network that encodes input functions or parameters (e.g., geometric features, initial or boundary conditions), and a trunk network that processes spatial and/or temporal coordinates.

The key innovation in Fusion DeepONet lies in its multi-scale feature fusion mechanism, which conditions each hidden layer of the trunk network on the corresponding hidden layer of the branch network. Let \( \mathbf{S}^{(l)}_B \in \mathbb{R}^{w} \) denote the cumulative feature vector obtained by summing all hidden features from the first to the \( (l-1) \)-th layer of the branch network, and let \( \mathbf{a}^{(l)}_T \in \mathbb{R}^{w} \) be the hidden feature vector at the \( l \)-th layer of the trunk network. The updated trunk representation at layer \( l \) is given by:
\begin{equation}
\mathbf{a}^{(l)}_T = \mathbf{S}^{(l)}_B \odot \sigma \left( \bm{W}_T^{(l)} \mathbf{a}^{(l-1)}_T + \mathbf{b}^{(l)} \right),
\label{eq:fusion_hidden}
\end{equation}
where \( \bm{W}_T^{(l)} \in \mathbb{R}^{w \times w} \) is a weight matrix, \( \mathbf{b}^{(l)} \in \mathbb{R}^{w} \) is a bias vector, \( \sigma \) is a nonlinear activation function, and \( \odot \) denotes element-wise (Hadamard) multiplication.

This layer-wise fusion allows the trunk network to adaptively incorporate geometry-dependent information at multiple scales, enhancing its expressivity in modeling complex spatio-temporal behavior—particularly in regions with high gradients or discontinuities.

Similar to the vanilla DeepONet, the final output is obtained via a contraction (inner product) between the outputs of the branch and trunk networks:
\begin{equation}
\bm{u}(\xi) = \sum_{k=1}^{p} \tilde{\bm{b}}_k(\mu) \cdot \tilde{t}_k(\xi),
\label{eq:fusion_output}
\end{equation}
where \( \tilde{\bm{b}}_k(\mu) \in \mathbb{R}^d \) denotes the \( k \)-th component of the vector-valued output from the final layer of the branch network, evaluated at the input parameters \( \mu \) (e.g., geometric or boundary features), and \( \tilde{t}_k(\xi) \in \mathbb{R} \) represents the corresponding scalar basis function from the fused trunk network, evaluated at the spatial-temporal coordinate \( \xi \in \mathbb{R}^n \). For additional details on the Fusion DeepONet architecture, see~\cite{peyvan2025fusion}.

\begin{figure}[!tbh]
    \centering
    \includegraphics[width=0.99\textwidth]{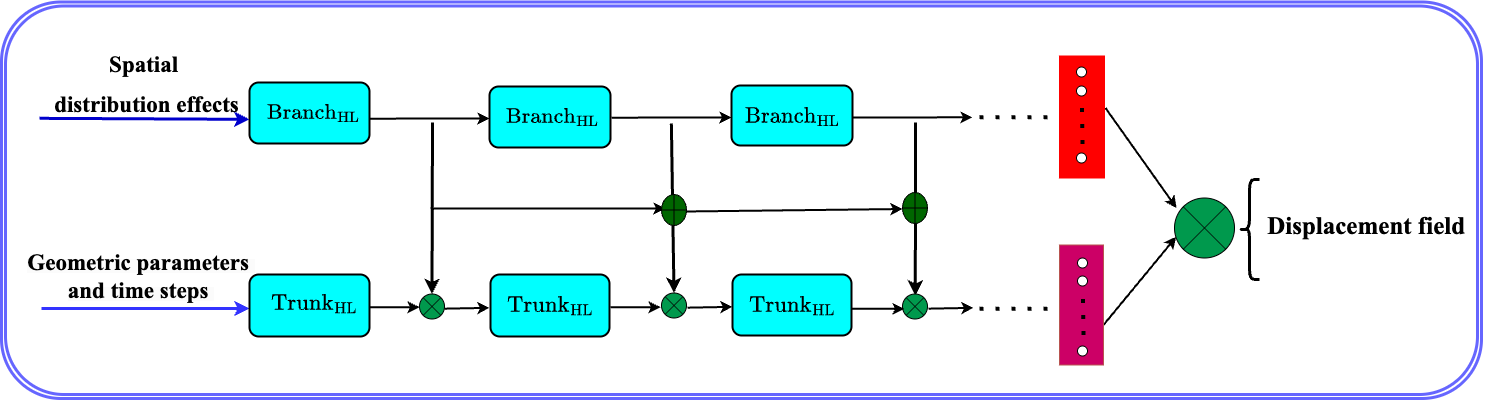}
    \caption{Schematic of the Fusion DeepONet architecture for displacement field prediction. The branch network encodes geometry-dependent input parameters such as the initial crack height \( h_i \) or hole radius \( r_i \). The trunk network takes as input the spatial coordinate \( \bm{x} \in \mathbb{R}^2 \) and time \( t \) . Each hidden layer of the trunk is conditioned on the corresponding hidden layer of the branch network (denoted as \texttt{Branch\textsubscript{HL}} and \texttt{Trunk\textsubscript{HL}}), while the final layer of the branch provides global conditioning to the output.}
    \label{fig:Fusion_Deeponet_Schematic}
\end{figure}

Figure~\ref{fig:Fusion_Deeponet_Schematic} illustrates the Fusion DeepONet architecture applied to displacement field prediction. In this formulation, the branch network captures input-dependent features such as variations in crack height \( h_i \) or hole radius \( r_i \), while the trunk network processes the query coordinates \( \bm{x} \in \mathbb{R}^2 \) and time \( t \). Unlike the standard DeepONet, the Fusion DeepONet incorporates hierarchical conditioning, where each hidden layer in the trunk is dynamically modulated by the corresponding hidden layer from the branch network. In addition, the final output of the branch network provides global conditioning at the output stage.

This layered interaction enables multi-scale feature fusion, allowing coarse geometric information to guide the learning of both low- and high-frequency components of the solution. As a result, the architecture is particularly effective at modeling solutions with sharp features or irregular geometries. To maintain consistent modulation across layers, the architecture assumes matching layer depths and widths in both branch and trunk networks, except for the final linear layer in the branch. Although this design enhances learning capacity and generalization, it introduces additional computational overhead compared to the vanilla DeepONet, especially during training.
\section{Performance of DeepONet and Fusion DeepONet}\label{sec:Performance of DeepONet and Fusion-DeepONet}
In this section, we compare the performance of the vanilla DeepONet and the Fusion-DeepONet in predicting crack paths across the three case studies listed in Table~\ref{tab:case_Cases}.
\begin{figure}[!tbh]
    \centering
    \begin{subfigure}{0.99\textwidth}
        \centering
        \includegraphics[width=\textwidth]{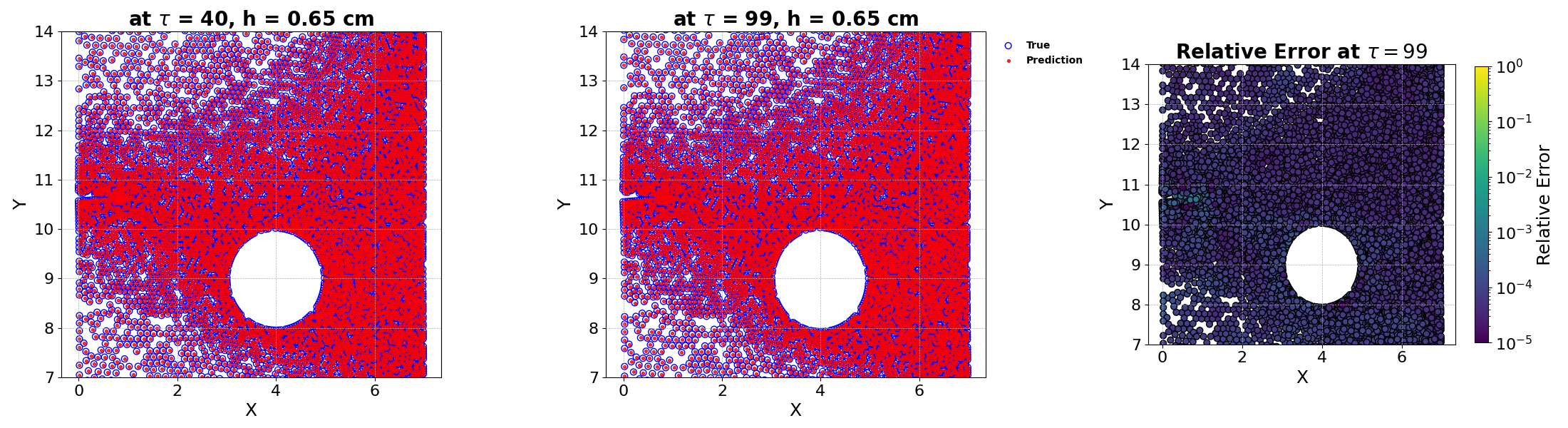}
        \caption{\textbf{Case 1 }: Prediction of particle positions using \textbf{Fusion DeepONet}.}
    \end{subfigure}

    \vspace{0.5em}
    \begin{subfigure}{0.99\textwidth}
        \centering
        \includegraphics[width=\textwidth]{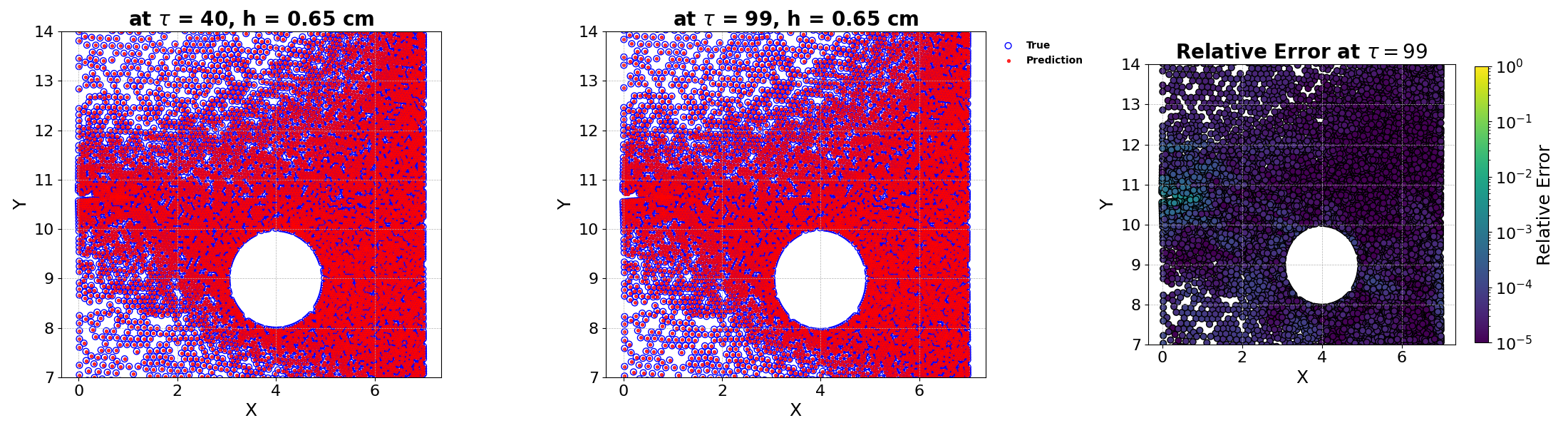}
        \caption{\textbf{Case 1 }: Prediction of particle positions using \textbf{vanilla DeepONet}.}
    \end{subfigure}
    \caption{
    Comparison of predicted and true particle positions at two time steps, \( \tau = 40 \) and \( \tau = 99 \), for \textbf{Case 1}, using (a) Fusion DeepONet and (b) vanilla DeepONet architectures. 
    Red points represent the predicted particle positions, while hollow blue circles denote the true positions. 
    The figure on the right shows the relative error at \( \tau = 99 \), comparing the predicted and true particle positions.
    }
    \label{fig:Fusion_prediction_H_nofail}
\end{figure}

\begin{figure}[!tbh]
    \centering
    \includegraphics[width=0.7\textwidth]{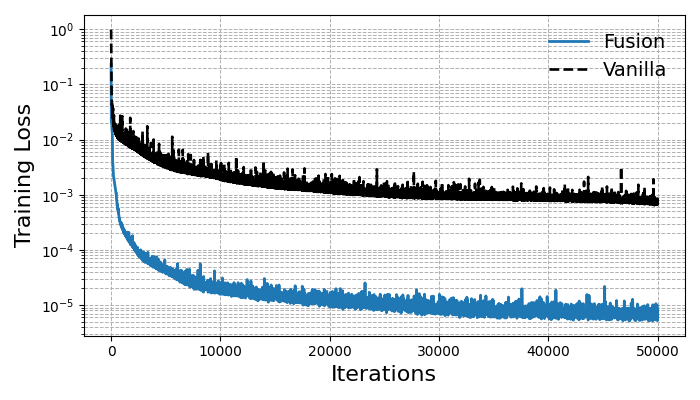}
    \caption{\textbf{Case 1: } Comparison of training loss between the Fusion DeepONet and the vanilla DeepONet over 50000 iterations.}
    \label{fig:loss_comparison_nofail}
\end{figure}

Figure~\ref{fig:Fusion_prediction_H_nofail} illustrates the predictive performance of (a) the Fusion DeepONet and (b) the vanilla DeepONet for \textbf{Case 1}, comparing predicted and true particle positions at two representative time steps: \( \tau = 40 \) and \( \tau = 99 \). The panel on the right shows the relative error at \( \tau = 99 \) for both architectures. In \textbf{Case 1}, the pre-crack notch height is varied while the hole radius is fixed at \( r = 1~\text{cm} \) (see Table~\ref{tab:case_Cases}). The dataset contains 40 samples, split into 35 for training and 5 for testing. Since no elements fail in this scenario, particles experience only incremental displacements over time without any fracture events.

The vanilla DeepONet architecture employs five hidden layers in both the branch and trunk networks, each with 100 neurons, and a latent dimension of 200 for the final basis projection. The \texttt{tanh} activation function is used throughout, and training is performed using the Adam optimizer from the \texttt{optax} library with a fixed learning rate of \( 1 \times 10^{-4} \).
In contrast, the Fusion DeepONet consists of three hidden layers in both subnetworks, each with 64 neurons, and a final latent layer also of size $64$ for trunk subnetwork and $2\times64$ for branch subnetwork. It utilizes the Rowdy activation function~\cite{jagtap2022deep}, a modified form of \texttt{tanh} designed to enhance expressive feature learning. The model is trained using the Adam optimizer with an exponentially decaying learning rate, starting from \( 1 \times 10^{-3} \), with a decay step of 2000 and a decay rate of 0.91.

In Figure~\ref{fig:Fusion_prediction_H_nofail}, red dots indicate predicted particle positions, while hollow blue circles denote the corresponding ground truth. Each red-blue pair corresponds to the same particle at a given time step, and their overlap visually reflects the prediction accuracy. The comparison demonstrates that both models can learn the particle dynamics, with Fusion DeepONet showing visibly improved alignment.
Figure~\ref{fig:loss_comparison_nofail} shows the evolution of training loss, measured as mean squared error (MSE), for both models. The Fusion DeepONet exhibits faster and more stable convergence, achieving an MSE as low as \( 10^{-5} \), compared to the vanilla DeepONet, which plateaus around \( 10^{-3} \).

\begin{figure}[!tbh]
    \centering
    \begin{subfigure}{0.99\textwidth}
        \centering
        \includegraphics[width=\textwidth]{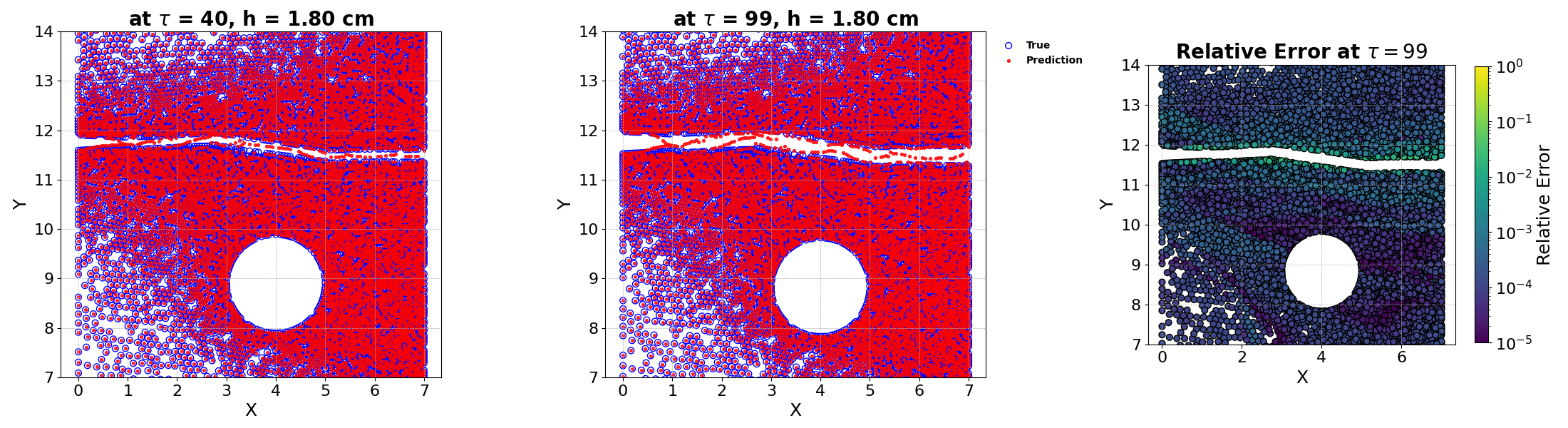}
        \caption{\textbf{Case 2: } Comparison between predicted and true crack path using \textbf{Fusion DeepONet}.}
    \end{subfigure}

    \vspace{0.5em}
    \begin{subfigure}{0.99\textwidth}
        \centering
        \includegraphics[width=\textwidth]{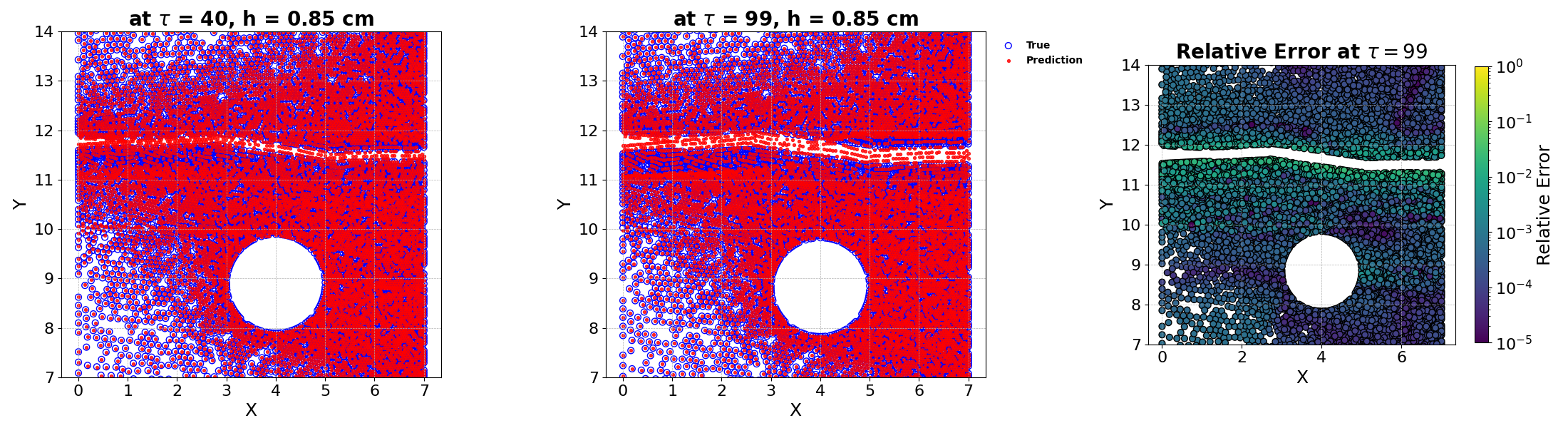}
        \caption{\textbf{Case 2: } Comparison between predicted and true crack path using \textbf{vanilla DeepONet}.}
    \end{subfigure}
    \caption{
    Predicted and true crack paths at two time steps, \( \tau = 40 \) and \( \tau = 99 \), along with the relative error at \( \tau = 99 \), comparing particle-level predictions using (a) Fusion DeepONet and (b) vanilla DeepONet architectures for \textbf{Case 2}. 
    Red points denote predicted particle positions, while hollow blue circles represent the corresponding ground truth. 
    Closer alignment between red and blue markers indicates higher predictive accuracy. }
    \label{fig:Fusion_prediction_H}
\end{figure}

\begin{figure}[!tbh]
    \centering
    \begin{subfigure}{0.99\textwidth}
        \centering
        \includegraphics[width=\textwidth]{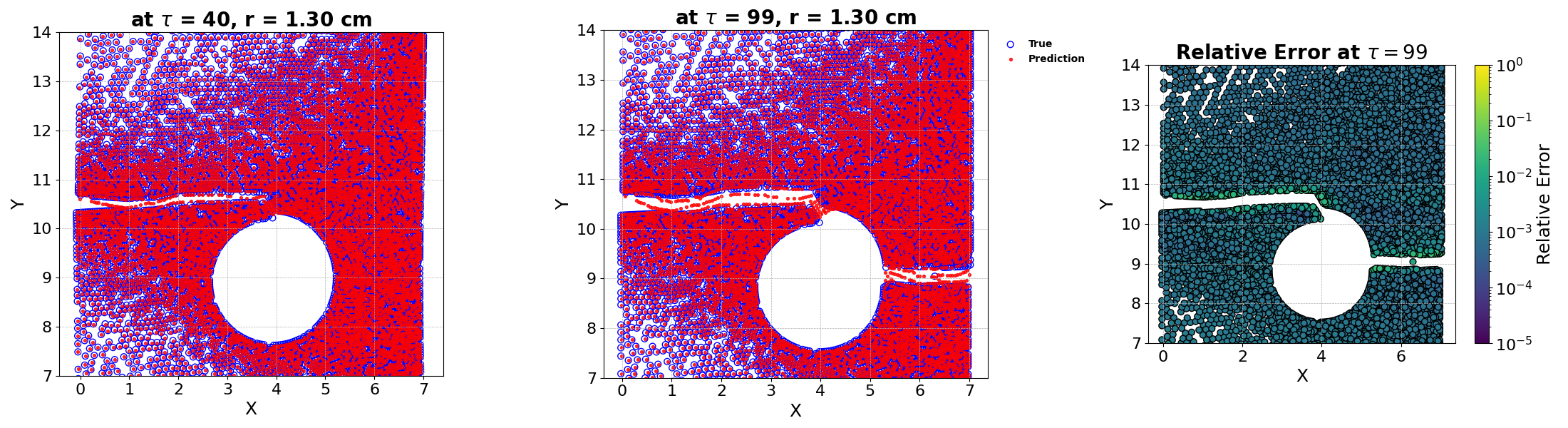}
        \caption{\textbf{Case 3: } Comparison between predicted and true crack path using \textbf{Fusion DeepONet}.}
    \end{subfigure}

    \vspace{0.5em}
    \begin{subfigure}{0.99\textwidth}
        \centering
        \includegraphics[width=\textwidth]{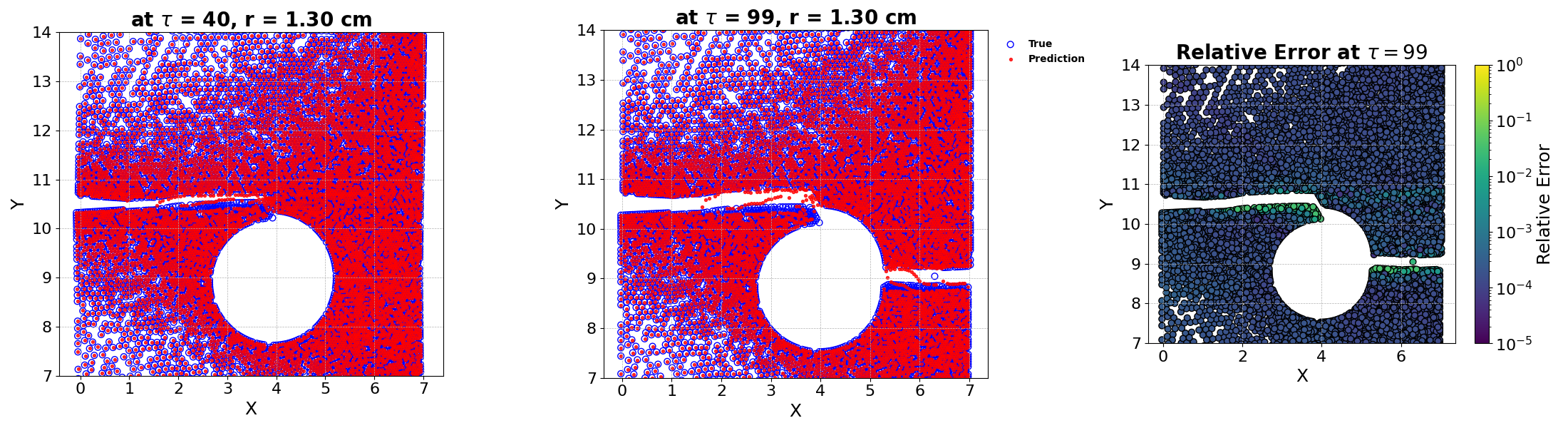}
        \caption{\textbf{Case 3: } Comparison between predicted and true crack path using \textbf{vanilla DeepONet}.}
    \end{subfigure}
\caption{Predicted and true crack paths at two time steps, \( \tau = 40 \) and \( \tau = 99 \), for \textbf{Case 3}, where the initial radius of the hole is varied. 
Red points indicate the predicted particle positions, while hollow blue circles represent the ground truth. 
Each red-blue pair corresponds to the same particle at a given time step. 
This comparison demonstrates the ability of both the Fusion and vanilla DeepONet architectures to generalize to unseen geometries. 
Notably, the Fusion DeepONet exhibits improved alignment with the true displacements, especially in regions where the crack path is highly sensitive to variations in hole radius.}
    \label{fig:Fusion_prediction_R}
\end{figure}


\begin{figure}[!tbh]
    \centering
    \begin{minipage}[b]{\textwidth}
        \centering
        \includegraphics[width=0.99\textwidth]{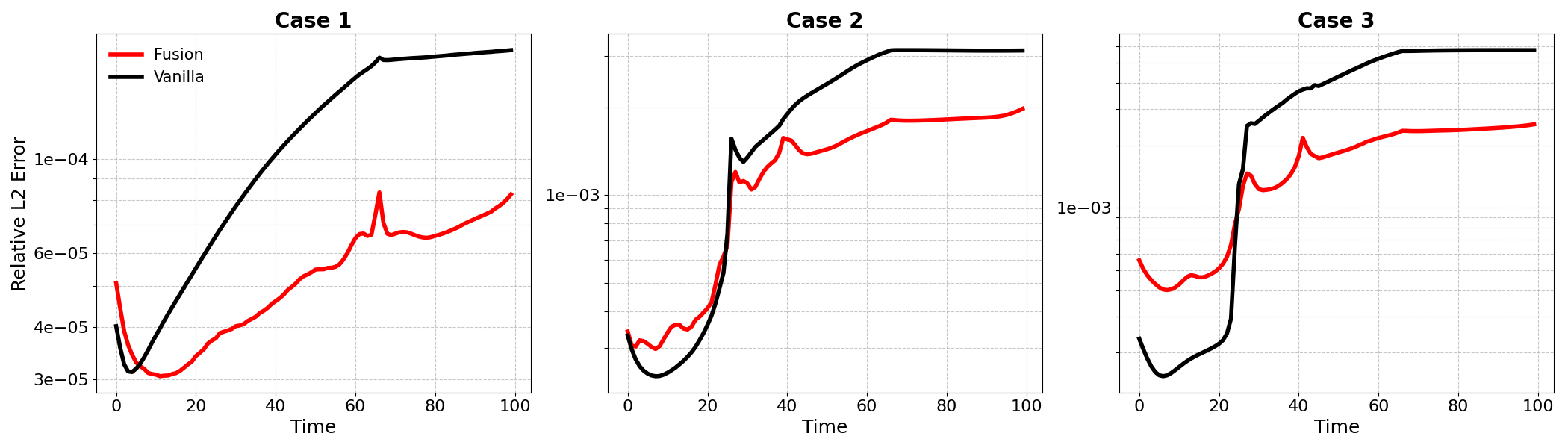}
        \caption*{\textbf{(a)} Relative $\mathcal{L}_2$ error over 100 time steps for \textbf{Case 1}, \textbf{Case 2}, and \textbf{Case 3}.}
    \end{minipage}
    \vspace{1em}
    \begin{minipage}[b]{\textwidth}
        \centering
        \includegraphics[width=0.99\textwidth]{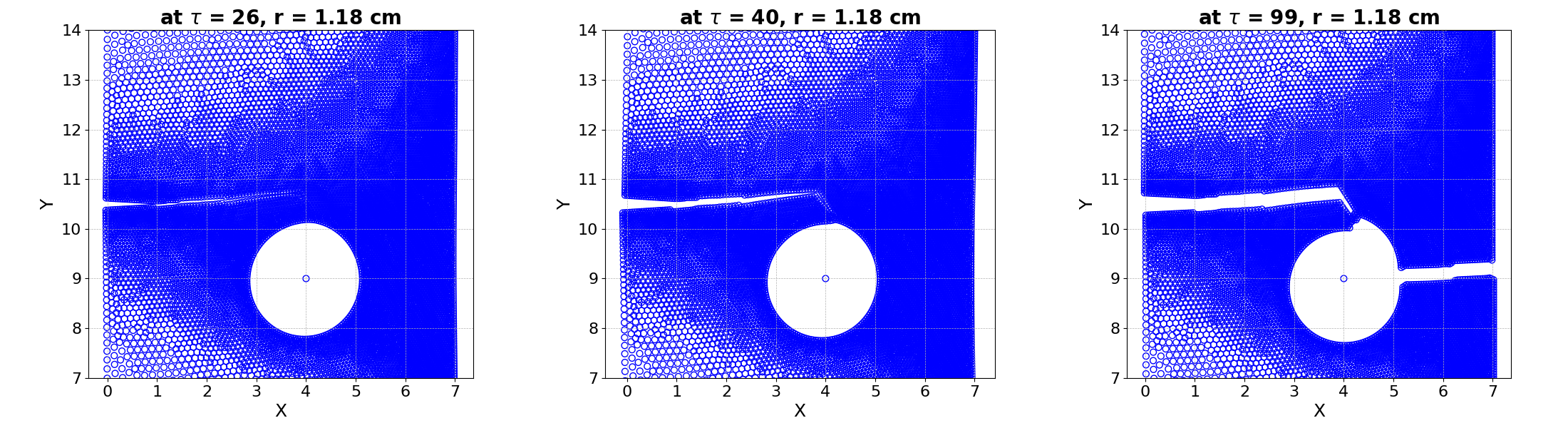}
        \caption*{\textbf{(b)} Crack propagation for \textbf{Case 3} at three selected time steps for $r = 1.18~\text{cm} $.}
    \end{minipage}
    \caption{Comparison of (a) the relative $\mathcal{L}_2$ error over time for three case studies reported in Table~\ref{tab:case_Cases}, and (b) crack propagation visualization for \textbf{Case 3} at three time steps when $r = 1.18~\text{cm} $.}
    \label{fig:l2_comparison}
\end{figure}

As summarized in Table~\ref{tab:case_Cases}, \textbf{Case 2} involves varying the pre-crack notch height while keeping the hole radius fixed at \( r = 1~\text{cm} \). The dataset consists of 50 samples with different notch heights, of which 45 are used for training and 5 for testing. In contrast, \textbf{Case 3} considers a fixed pre-crack notch height while varying the initial hole radius over the range \( r = 0.5~\text{cm} \) to \( r = 1.5~\text{cm} \). This dataset includes 51 samples, with 45 used for training and 6 for testing.
The same architectures used in \textbf{Case 1} are employed here for both the vanilla DeepONet and Fusion DeepONet models. Specifically, the vanilla DeepONet uses five hidden layers in both the branch and trunk networks, each containing 100 neurons, and a latent dimension of 200 for the final basis projection. The Fusion DeepONet consists of three hidden layers in both subnetworks, each with 64 neurons, and a final latent layer also of size 64 for trunk network and $2\times64$ for branch networks.

Figures~\ref{fig:Fusion_prediction_H} and~\ref{fig:Fusion_prediction_R} illustrate the performance of both models across 100 time steps for \textbf{Case 2} and \textbf{Case 3}, respectively. In each plot, predicted and true particle positions are compared at two representative time steps: \( \tau = 40 \) and \( \tau = 99 \). Red points represent predicted particle positions, while hollow blue circles denote the corresponding ground truth. Each red-blue pair corresponds to the same particle at a specific time, and the degree of overlap between the markers reflects the prediction accuracy.
These results demonstrate that both models effectively learn the particle dynamics and accurately predict the crack path. Notably, they capture the interaction between the crack and the hole, particularly in samples where the crack propagates near or intersects the hole boundary.
Figure~\ref{fig:Fusion_prediction_H} presents the predictions for a sample with \( h = 0.85~\text{cm} \), where the crack does not interact with the hole. In contrast, Figure~\ref{fig:Fusion_prediction_R} shows predictions for a sample with \( r = 1.30~\text{cm} \), where the crack interacts with the hole before failure occurs.
In both cases, the relative error ranges from \( 10^{-5} \) to \( 10^{0} \). As expected, the prediction error is notably higher in regions near the crack tip and path, while it remains lower in areas farther from the crack, indicating localized difficulty in capturing the highly nonlinear fracture behavior.


\begin{table}[ht]
\centering
\renewcommand{\arraystretch}{1.9}
\caption{Training time and relative error for vanilla and Fusion DeepONet.}
\begin{tabular}{|>{\centering\arraybackslash}p{3.2cm}|
                >{\centering\arraybackslash}p{3.5cm}|
                >{\centering\arraybackslash}p{3.5cm}|
                >{\centering\arraybackslash}p{3.5cm}|}
\hline
\textbf{Method} & \textbf{Case 1} & \textbf{Case 2} & \textbf{Case 3 } \\
\hline
Vanilla DeepONet (Time) & 1752 min & 3245 min & 2986 min \\
Fusion DeepONet (Time)  & 996 min  & 1294 min & 1143 min \\
\hline
Vanilla DeepONet (Error) & 6.20e-05 & 9.86e-04 & 8.46e-04 \\
Fusion DeepONet (Error)  & 4.97e-05 & 5.13e-04 & 1.01e-03 \\
\hline
\end{tabular}
\label{tab:training_time_error}
\end{table}

Figure~\ref{fig:l2_comparison}(a) presents a comparison of the relative $\mathcal{L}_2$ error over 100 time steps for all three cases, using both the vanilla DeepONet and the Fusion DeepONet. Across all scenarios, the Fusion DeepONet consistently outperforms the vanilla architecture, achieving lower prediction errors in modeling crack path evolution.
Among the three cases, \textbf{Case 1} proves to be the simplest to predict. This case involves only particle displacements over time, with no element failure, making it a pure displacement prediction problem. In contrast, \textbf{Case 2} and \textbf{Case 3} involve both displacement and fracture, significantly increasing the complexity of the prediction task due to the presence of discontinuities and topological changes.
Additionally, the plot reveals that the relative $\mathcal{L}_2$ error tends to increase over time, particularly in \textbf{Case 2} and \textbf{Case 3}. The observed spikes in error correspond to the onset of fracture events, while the models initially predict displacements accurately, the emergence of failed elements introduces sudden changes in the system's dynamics, making accurate prediction more difficult in later time steps.

In \textbf{Case 1}, the relative $\mathcal{L}_2$ error increases gradually over time and eventually stabilizes, reflecting a purely elastic response with no crack propagation. This stable behavior indicates that the model effectively captures the physics of the elastic regime. In contrast, \textbf{Case 2} and \textbf{Case 3} show a sharp increase in error after approximately $\tau = 20$ time steps. As illustrated in Figure~\ref{fig:l2_comparison}(b), this jump corresponds to the onset of crack propagation, where the solution begins to accountsfor material separation during fracture. 
A subsequent spike in error is observed around $\tau = 40$, coinciding with the moment the crack re-nucleates from the central hole in scenarios involving interaction with the hole. The onset of crack nucleation introduces geometric discontinuities and increases the complexity of the system, making the prediction task more challenging, especially for the vanilla DeepONet. The error continues to rise until approximately $\tau = 66$ after which the system is allowed to equilibriate without loading, after which it stabilizes, indicating that the system is reaching a new equilibrium state by the final time step ($\tau = 100$).

Table~\ref{tab:training_time_error} presents the training error and training time for displacement prediction in all three cases using the vanilla DeepONet and the Fusion DeepONet. The training error for \textbf{Case 1} is lower than that of the other two cases, which is expected given the simpler physical setting—no fracture—and the smaller number of training samples (35 samples for \textbf{Case 1}, compared to 45 samples for both \textbf{Case 2} and \textbf{Case 3}).

The training time for the vanilla DeepONet is higher than that of the Fusion DeepONet. This is primarily because the vanilla DeepONet was trained for 60,000 iterations, whereas the Fusion DeepONet was trained for 50,000 iterations. These iteration counts were chosen empirically based on convergence behavior; training was stopped once the loss plateaued and no significant improvement was observed.

In addition to differences in iteration counts, the architecture of the vanilla DeepONet is more complex. Both its branch and trunk networks consist of 5 hidden layers with 100 neurons per layer. In contrast, the Fusion DeepONet uses only 3 hidden layers with 64 neurons per layer in both networks. This reduced model size in the Fusion DeepONet results in a smaller computational footprint and faster training, despite achieving better predictive performance.

Table~\ref{tab:training_time_error} also summarizes the average relative $\mathcal{L}_2$ error for the vanilla and Fusion DeepONet methods across the three case studies. While the average errors for each case are relatively close between the two methods, a more detailed analysis in Figure~\ref{fig:l2_comparison}(a) reveals a key distinction. Specifically, the Fusion DeepONet consistently outperforms the vanilla DeepONet after $\tau = 20$, coinciding with the onset of crack propagation. This performance advantage becomes especially apparent as the system transitions from an elastic to a fractured state, where predictive accuracy becomes more challenging. The comparable average errors reported in the table arise because the early elastic regime dominates the error average, despite the superior performance of Fusion in the later fracture-dominated phase.

\section{Summary}\label{sec:Summary}

We investigated the application of two variants of DeepONet—vanilla and Fusion—for modeling crack propagation in specimens with varying geometries. The training data for these models was obtained from Constitutively Informed Particle Dynamics (CPD) simulations, which provide a physics-informed framework for capturing both crack nucleation and propagation. The networks were trained using inputs such as the pre-crack notch height and hole radius, which varied across different samples. In \textbf{Case~1}, only the notch height was varied, and no element failure occurred, making it a pure displacement prediction task. In \textbf{Cases~2} and \textbf{3}, the notch height and the hole radius were varied, respectively, and fracture evolution was included by accounting for element failure. The number of training samples used in each case was 32 (varying notch heights), 45 (varying notch heights), and 45 (varying radii), respectively. In all cases, the trunk network received spatial coordinates and time over 100 time steps as input.

The Fusion DeepONet achieved lower prediction errors in modeling the spatiotemporal evolution of displacements and cracks. Among the three cases, \textbf{Case~1} exhibited the lowest overall error, as it involves only elastic displacement without crack propagation. The relative $\mathcal{L}_2$ error in this case increases gradually and eventually plateaus, indicating stable and accurate predictions throughout the elastic regime.
In contrast, \textbf{Cases~2} and \textbf{3} involve fracture dynamics, making them significantly more challenging due to the presence of topological discontinuities and rapidly evolving damage. Both cases exhibit a sharp rise in relative error during the onset of crack initiation. Before this point, the networks accurately capture the displacement fields. However, as elements begin to fail, abrupt transitions caused by material separation introduce significant challenges for prediction. These spikes in error closely coincide with fracture nucleation events.

This study underscores the potential of integrating the CPD simulation framework with DeepONet-based operator learning to model complex fracture behavior in discontinuous domains. CPD's ability to capture crack nucleation and evolution—without the limitations of continuum assumptions—provides high-fidelity training data, while the Fusion DeepONet architecture enhances prediction accuracy and generalization. This hybrid approach offers a powerful and efficient surrogate modeling strategy for accelerating fracture simulations across a broad range of geometries and loading conditions.

\section*{Acknowledgments}
This research was primarily supported as part of the AIM for Composites, an Energy Frontier Research Center funded by the U.S. Department of Energy (DOE), Office of Science, Basic Energy Sciences (BES), under Award \#DE-SC0023389 (computational studies, data analysis). 
Computing facilities were provided by the Center for Computation and Visualization, Brown University. 

\clearpage
\bibliographystyle{unsrt}  
\bibliography{sample}  

\end{document}